\documentclass{article}

    \PassOptionsToPackage{numbers, sort&compress}{natbib}

\usepackage[main,final]{neurips_2026}

\usepackage[utf8]{inputenc} %
\usepackage[T1]{fontenc}    %
\usepackage{hyperref}       %
\usepackage{url}            %
\usepackage{booktabs}       %
\usepackage{amsfonts}       %
\usepackage{nicefrac}       %
\usepackage{microtype}      %
\usepackage{xcolor}         %
\usepackage{graphicx}       %
\usepackage{subcaption}     %
\usepackage{multirow}       %
\usepackage{algorithm}      %
\usepackage{algpseudocode}  %
\usepackage{amsmath}
\usepackage{amsthm}
\let\AnoMixMakeTitle\maketitle
\usepackage{authblk}
\usepackage{latexml}

\let\maketitle\AnoMixMakeTitle

\theoremstyle{definition}

\title{Distilling Multi-Teacher Scoring Principles for Unsupervised Time Series Anomaly Detection}

\iflatexml
\author{\textbf{Dongchan Cho}\textsuperscript{*}}
\affil{Industrial AI Lab, SimPlatform Co. Ltd. Affiliate Research Institute\\
    \href{mailto:cdc0213@simplatform.com}{\texttt{cdc0213@simplatform.com}}\\
    \textsuperscript{*}Corresponding author.}
\author{\textbf{Jiho Han}}
\affil{Industrial AI Lab, SimPlatform Co. Ltd. Affiliate Research Institute\\
    \href{mailto:jihohan@simplatform.com}{\texttt{jihohan@simplatform.com}}}
\author{\textbf{Juwon Hwang}}
\affil{Industrial AI Lab, SimPlatform Co. Ltd. Affiliate Research Institute\\
    \href{mailto:jwhwang@simplatform.com}{\texttt{jwhwang@simplatform.com}}}
\author{\textbf{Changhee Min}}
\affil{Industrial AI Lab, SimPlatform Co. Ltd. Affiliate Research Institute\\
    \href{mailto:chmin@simplatform.com}{\texttt{chmin@simplatform.com}}}
\author{\textbf{Minsang Kim}}
\affil{Industrial AI Lab, SimPlatform Co. Ltd. Affiliate Research Institute\\
    \href{mailto:mskim@simplatform.com}{\texttt{mskim@simplatform.com}}}
\author{\textbf{Namsoon Jung}}
\affil{Industrial AI Lab, SimPlatform Co. Ltd. Affiliate Research Institute\\
    \href{mailto:nsj@simplatform.com}{\texttt{nsj@simplatform.com}}}
\else
\author{Dongchan Cho\thanks{Corresponding author: Dongchan Cho.}}
\author{Jiho Han}
\author{Juwon Hwang\authorcr}
\author{Changhee Min}
\author{Minsang Kim}
\author{Namsoon Jung}
\affil{Industrial AI Lab, SimPlatform Co. Ltd. Affiliate Research Institute\\
    \texttt{\{cdc0213, jihohan, jwhwang, chmin, mskim, nsj\}@simplatform.com}}
\fi

\hypersetup{%
    pdfauthor={Dongchan Cho, Jiho Han, Juwon Hwang, Changhee Min, Minsang Kim, Namsoon Jung}
}

\begin{document}

\maketitle

\begin{abstract}
Time series anomalies take diverse forms, and individual detectors encode different anomaly scoring principles that rarely cover all anomaly types.
Recent unsupervised time-series anomaly detectors provide complementary scoring signals, yet deploying them as an ensemble preserves the inference and maintenance cost of every individual detector.
We study whether heterogeneous teacher detectors can be distilled into a single lightweight detector in the unsupervised setting, where no anomaly labels are available to define a shared target.
\textbf{AnoMix} addresses this by extracting weak pairwise supervision from the training-set anomaly scores of frozen teachers.
It normalizes each teacher score sequence independently and trains the student with CoRank, a multi-teacher ranking-distillation objective that converts teacher score gaps into pairwise supervision through teacher consensus.
Across seven benchmark datasets, AnoMix successfully integrates complementary teacher scoring principles into a single deployable model, improving over individual detectors and static score ensembles while using a compact student with under 80k parameters on the benchmark settings.
A zero-shot variant further shows that the distilled ranking signal transfers to unseen datasets and can outperform large pretrained time-series foundation models for anomaly detection.
\end{abstract}

\section{Introduction}
\label{sec:introduction}

Time series anomaly detection (TSAD) is a critical problem in industrial monitoring~\citep{PSM-2021}, cyber-physical systems~\citep{SWaT-2016}, and sensor analytics~\citep{OmniAnomaly-2019, SMAP_MSL-2018, Daphnet-2010}.
Because anomalies are open-set events and anomaly labels are scarce, TSAD is often addressed in an unsupervised setting.
In this setting, the anomaly criterion is ambiguous, and a detector must infer which temporal deviations should be treated as abnormal from the training series alone.
Recent benchmark studies~\citep{Elephant_in_the_Room-2024, TAB-2025, QuoVadisTAD-2024} highlight that this ambiguity leads different detectors to define and detect anomalies under their own criteria, with no single detector dominating across datasets.
Together, these observations suggest that improving TSAD performance requires using multiple anomaly-scoring principles.

Existing TSAD detectors already provide a rich source of such complementary scoring principles.
Reconstruction-based models emphasize reconstruction failure~\citep{USAD-2020, OmniAnomaly-2019}, forecasting models emphasize predictive residuals~\citep{TranAD-2022, TimesNet-2022}, latent-space models emphasize deviation from compact normal representations~\citep{OracleAD-2025, DeepSVDD-2018}, and association-based models emphasize structural and temporal dependency mismatch~\citep{CrossAD-2025, AnomalyTransformer-2021}.
Because their scores reflect different abnormality criteria, these detectors can complement one another~\citep{Aggarwal-2015-ensembles}.
A direct way to exploit this complementarity is to deploy them together and combine their scores.
However, such ensembles preserve the inference cost and maintenance overhead of all individual models~\citep{LSCP-2019, SUOD-2021, MetaOD-2021}.
A more practical goal is to transfer the complementary scoring principles into one lightweight student.

\emph{Knowledge distillation} appears to be a natural route toward this goal because teachers can be used during training while only the student is deployed at inference~\citep{MultiTeacher-KD-2020, Hinton-KD-2015}.
However, applying distillation to unsupervised TSAD raises a more fundamental question than how to compress a model: what exactly should be distilled?
In supervised classification, teacher and student models predict probabilities over the same set of classes, and ground-truth labels anchor the meaning of the teacher signal~\citep{Hinton-KD-2015}.
In unsupervised TSAD, neither condition holds.
Although each teacher outputs anomaly scores, the scores are continuous and detector-specific, and the label-free setting provides no direct way to verify whether they are correct.
When multiple teachers are used, this mismatch becomes more pronounced because their numerical magnitudes reflect different abnormality criteria and different calibrations.
Therefore, direct score regression is structurally mismatched with this setting, as it asks one student to fit multiple teacher-specific score scales as if they formed a single consistent target.

Our key insight is that imperfect teacher scores can still provide useful weak supervision through pairwise anomaly comparisons rather than through pointwise regression targets.
Although two teachers may use different score scales and abnormality criteria, they can still agree that one time point appears more anomalous than another.
We call this relation a pairwise \emph{anomaly preference}.
Such preferences provide a more stable supervision form than pointwise regression targets because they compare samples within each teacher score sequence instead of requiring a shared calibrated score scale across teachers.
Moreover, when multiple heterogeneous teachers support the same pairwise direction, their agreement suggests that the comparison is informative.
Thus, unsupervised TSAD distillation becomes feasible by training a single student to learn reliable pairwise anomaly preferences from multi-teacher score sequences computed on the training data.

Building on this insight, we introduce \textbf{AnoMix}, a label-free multi-teacher distillation framework for learning
a neural anomaly-scoring function. Unlike conventional TSAD models that derive
anomaly scores from reconstruction, forecasting, association, or representation discrepancies,
AnoMix trains a student to directly output the anomaly score used at inference.
Since no labeled anomaly-score target exists, AnoMix supervises this scalar score through
pairwise anomaly preferences derived from frozen teacher score sequences.
First, AnoMix normalizes each frozen teacher score sequence independently and then trains a compact student using \emph{CoRank}, a consensus-ranking objective over mini-batch time-point pairs.
CoRank constructs consensus support from teacher-level score gaps, assigns a consensus direction induced by normalized teacher differences, and weights each comparison by consensus confidence.
The student is therefore not asked to reproduce teacher-specific pointwise score magnitudes.
Instead, it learns a neural anomaly-scoring function aligned with the teacher-consensus ordering induced by CoRank.

Our contributions are as follows:
\begin{itemize}
    \item We formulate label-free multi-teacher TSAD distillation as direct anomaly-score prediction for deployment, where a lightweight student learns a neural anomaly-scoring function from heterogeneous unsupervised teacher score sequences.

    \item We propose AnoMix, which turns normalized teacher score sequences computed on the training data into pairwise anomaly-preference supervision, allowing a single student to learn anomaly-scoring principles without regressing to teacher-specific pointwise scores.

    \item Across TSAD benchmark datasets and the zero-shot cross-dataset setting, AnoMix and AnoMix-ZS achieve strong performance against individual detectors, static score-ensemble baselines, and pretrained foundation-model baselines, demonstrating that pairwise anomaly preferences from multiple teachers can be distilled into a single lightweight student.
\end{itemize}

\section{Related Work}

\paragraph{Knowledge Distillation.}
Knowledge distillation (KD) was originally introduced as a way to compress a large teacher into a compact student by matching softened output distributions~\citep{Hinton-KD-2015}.
Later KD methods broadened the transferred signal from logits to intermediate features~\citep{FitNets-2015}, contrastive representations~\citep{CRD-2020}, and sample relations such as activation similarities, similarity rankings, and relational structures~\citep{SPKD-2019,DarkRank-2018,Park_2019_CVPR}.
Multi-teacher KD further aggregates supervision from several teachers into a single student~\citep{MultiTeacher-KD-2020}.
Together, these lines show that KD need not be limited to output matching, but they still typically assume a well-defined transfer target.
For anomaly detection, distillation has been explored primarily in the image domain, often with a single teacher and with teacher--student discrepancy itself serving as the anomaly signal~\citep{KD_ImageAD_STPM-2021, KD_ImageAD_MKD-2021}.
In time-series anomaly detection, distillation remains relatively underexplored.
Recent work such as AnomalyLLM studies single-teacher transfer through feature mimicry and discrepancy-based detection~\citep{AnomalyLLM-2024}.
However, prior work does not address the AnoMix setting of label-free transfer from heterogeneous TSAD teachers to a single student.
\vspace{-0.5em}

\paragraph{Learning-to-Rank Methods.}
Learning-to-rank methods learn scoring functions from relative ordering constraints rather than calibrated target values.
Existing ranking objectives, however, typically assume that the target ordering is already available.
Pairwise methods such as RankNet~\citep{RankNet-2005} optimize pairwise preferences or teacher-induced orderings, max-margin ranking losses~\citep{LearningToRank-2009} impose ordered margin constraints, and listwise methods such as ListNet~\citep{ListNet-2007} optimize an ordering distribution over a set of samples.
More recent differentiable sorting and ranking surrogates~\citep{Blondel-2020} further make rank-based objectives compatible with gradient optimization.
Relation-based distillation similarly transfers sample-level structure rather than only pointwise outputs~\citep{SPKD-2019,DarkRank-2018,Park_2019_CVPR}.
These methods primarily specify how to optimize a given target ordering or relation.
AnoMix addresses a different setting.
In our setting, the target relation is not given by labels or by a single calibrated teacher.
AnoMix must construct pairwise supervision from heterogeneous teacher score sequences.
\vspace{-0.5em}

\begin{figure}[t]
    \centering
    \includegraphics[width=\textwidth]{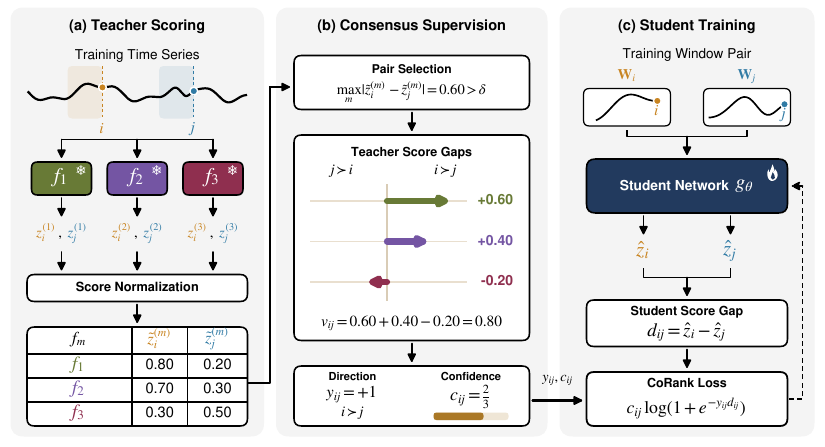}
    \caption{Training overview of AnoMix.
    (a) Frozen teachers (\protect\raisebox{-0.12em}{\protect\includegraphics[height=1em]{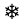}}) score the unlabeled training series, followed by score normalization.
    (b) CoRank selects informative pairs and derives their consensus direction and confidence.
    (c) A trainable student (\protect\raisebox{-0.12em}{\protect\includegraphics[height=1em]{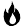}}) learns the pairwise ordering through a confidence-weighted loss.}
    \label{fig:architecture}
\vspace{-1em}
\end{figure}

\section{Methodology}
\label{sec:methodology}
\vspace{-0.5em}
Figure~\ref{fig:architecture} follows a concrete pair of endpoints through AnoMix.
In panel~(a), each teacher $f_m$ produces raw anomaly scores $z_i^{(m)},z_j^{(m)}$ aligned with endpoints $i,j$ of the unlabeled training series, with its parameters fixed throughout student training.
Per-teacher min--max normalization using training statistics yields the scores $\tilde z_i^{(m)},\tilde z_j^{(m)}$ in row $m$ of the figure.
In panel~(b), CoRank retains the pair if any teacher provides a sufficiently large score gap, combines the signed gaps into a consensus direction, and measures their agreement as a confidence weight.
The illustrated pair remains informative despite one opposing teacher: its consensus favors $i$ over $j$ with confidence $2/3$.
In panel~(c), the same lightweight student processes both windows and learns this preference through a confidence-weighted pairwise loss.
At inference, all teachers are discarded, leaving only the student to produce pointwise anomaly scores.
A detailed rationale for CoRank-based multi-teacher ranking distillation is provided in Appendix~\ref{app:theory}.
\vspace{-0.5em}
\subsection{Problem Setup}
\vspace{-0.5em}
Let $\mathbf{X}_{\mathrm{train}} = (\mathbf{x}_1,\dots,\mathbf{x}_T) \in \mathbb{R}^{T \times D}$ denote an unlabeled multivariate training time series.
For each endpoint $t \ge L$, we form a sliding-window sample $\mathbf{W}_t = (\mathbf{x}_{t-L+1}, \dots, \mathbf{x}_t) \in \mathbb{R}^{L \times D}$.
We assume access to $M$ frozen pre-trained teacher detectors $\{f_1,\dots,f_M\}$.
Each teacher provides an anomaly score $z_t^{(m)}$ aligned with endpoint $t$, and these teacher scores are normalized independently.
The student takes the sliding-window sample $\mathbf{W}_t$ as input and outputs a scalar prediction $\hat{z}_t$ for the same endpoint.
No anomaly labels are used, and the normalized teacher scores provide the only supervision signal for training.
After training, the teacher detectors are discarded, and the student alone processes each test window to produce a scalar anomaly score.
The goal is to learn a single deployable student from unlabeled data and frozen teacher score sequences by capturing pairwise anomaly preferences supported by multiple teachers.
\vspace{-0.5em}

\subsection{Teacher Score Normalization}
\vspace{-0.5em}
Each teacher produces anomaly scores on a detector-specific scale.
We therefore min-max normalize each teacher score sequence independently using only training-set statistics.
This yields normalized teacher scores $\tilde{z}_t^{(m)} \in [0,1]$ on the training series.
This preserves within-teacher ordering while reducing cross-teacher scale mismatch for distillation.
Although teacher score normalization does not make teacher scores a reliable shared regression target, it makes teacher-specific score gaps comparable enough for pairwise supervision because their possible ranges are aligned.
AnoMix then computes pairwise teacher score gaps from these normalized scores.
These gap magnitudes are used later to select informative pairs and weight pairwise comparisons derived from teacher scores.

\vspace{-0.5em}

\subsection{Student Architecture}
\label{sec:student_arch}
\vspace{-0.5em}
We use a lightweight causal temporal convolutional network (TCN) student with one scalar output.
This architecture matches the endpoint-based scoring protocol, where each input window is assigned the anomaly score of its final timestep.
Given a window $\mathbf{W}_t$, a linear projection maps it to $\mathbf{H}^{(0)} \in \mathbb{R}^{L \times H}$, $N$ causal temporal blocks~\citep{TCN-2018} produce $\mathbf{H}^{(N)}$, and the endpoint representation $\mathbf{h}_t = \mathbf{H}^{(N)}_L \in \mathbb{R}^{H}$ is mapped to a scalar score $\hat{z}_t = \mathbf{w}^{\top}\mathbf{h}_t + b$.
This lightweight design helps separate the effect of AnoMix's multi-teacher ranking distillation from gains due to a complex student architecture.
Architectural details for both the main student and the zero-shot variant are provided in Appendix~\ref{app:student_architectures}.

\vspace{-0.5em}
\subsection{CoRank Loss}
\vspace{-0.5em}
The main challenge is not choosing a ranking loss, but constructing reliable ranking supervision from heterogeneous teachers.
CoRank therefore uses a standard pairwise logistic form, but differs from conventional ranking losses in the situation it addresses.
RankNet~\citep{RankNet-2005}, max-margin ranking~\citep{LearningToRank-2009}, listwise ranking~\citep{ListNet-2007}, and differentiable sorting and ranking objectives~\citep{Blondel-2020} can be considered when a target ordering is already available.
In AnoMix, no anomaly labels are available, and no single teacher provides a reliable target ordering on its own.
A straightforward baseline is to regress student outputs directly to teacher scores.
In TSAD, however, direct regression can spend student capacity fitting the score scale of many normal or low-score regions, and detector-specific calibration differences remain even after normalization.
AnoMix instead learns from pairwise score gaps whose directions are supported by multiple teachers.

Consider a mini-batch of $B$ windows.
Let $\tilde{z}_i^{(m)}$ denote the normalized score assigned by teacher $m$ to sample $i$, and let $\hat{z}_i$ denote the corresponding student prediction.
We denote teacher $m$'s signed score gap by $\Delta_{ij}^{(m)} = \tilde{z}_i^{(m)} - \tilde{z}_j^{(m)}$.
From the batch, we enumerate all unordered non-identical pairs $(i,j)$ with $i<j$ and retain only those pairs for which at least one teacher exhibits a meaningful score gap:
\begin{equation}
    \mathcal{P}
    =
    \left\{
    (i,j)\;:\; 1 \le i < j \le B,\;
    \max_m \left|\Delta_{ij}^{(m)}\right| > \delta
    \right\}.
\end{equation}
The pair selection step in panel~(b) of Figure~\ref{fig:architecture} removes pairs whose teacher score gaps are too small to provide meaningful anomaly-preference supervision.
The resulting set $\mathcal{P}$ defines the \emph{consensus support} and contains the meaningful teacher-induced comparisons used for consensus supervision before teacher scores are averaged.

Once a pair $(i,j)$ is retained in $\mathcal{P}$, every teacher contributes its signed score gap $\Delta_{ij}^{(m)}$.
Using all teacher score gaps for each retained pair allows the final supervision to reflect the full degree of cross-teacher agreement or disagreement on that pair, rather than only the strongest response.
We aggregate these signed gaps to form the teacher consensus
\begin{equation}
    v_{ij}
    =
    \sum_{m=1}^{M}
    \Delta_{ij}^{(m)},
    \qquad
    c_{ij}
    =
    \frac{|v_{ij}|}
    {\sum_{m=1}^{M}\left|\Delta_{ij}^{(m)}\right|}.
\end{equation}
Here, $y_{ij} = \mathrm{sgn}(v_{ij}) \in \{-1,0,1\}$ defines the consensus direction, where $\mathrm{sgn}(\cdot)$ is the sign function: $y_{ij}=1$ if $v_{ij}>0$, $y_{ij}=-1$ if $v_{ij}<0$, and $y_{ij}=0$ if $v_{ij}=0$.
The quantity $c_{ij}\in[0,1]$ quantifies how consistently teachers support that direction, as visualized by the confidence bar in Figure~\ref{fig:architecture}.
When all nonzero signed gaps point in the same direction, $c_{ij}=1$.
When their signed score gaps strongly conflict, $c_{ij}$ approaches $0$.
A high confidence weight indicates agreement among teacher score gaps, not correctness of the resulting preference.
If $v_{ij}=0$, then $y_{ij}=0$, $c_{ij}=0$, and the pair contributes no effective supervision.

The student's pairwise anomaly preference is represented by the score margin $d_{ij} = \hat{z}_i - \hat{z}_j$, the signed difference between student predictions for samples $i$ and $j$.
The CoRank loss in panel~(c) of Figure~\ref{fig:architecture} penalizes student margins $d_{ij}$ that disagree with the teacher consensus direction $y_{ij}$, with each pair weighted by its consensus confidence $c_{ij}$:
\begin{equation}
    \mathcal{L}_{\mathrm{CoRank}}
    =
    \frac{1}{|\mathcal{P}|}
    \sum_{(i,j)\in\mathcal{P}}
    c_{ij}
    \log\!\Big(
        1 + \exp\!\big(
            - y_{ij} d_{ij}
        \big)
    \Big).
\end{equation}
If no valid pair is retained in a mini-batch, we define the batch loss as zero and skip its contribution.

CoRank can be viewed as confidence-weighted distillation of pairwise anomaly preferences induced by normalized teacher scores.
It avoids fitting teacher-specific pointwise scores and does not construct pairwise supervision from an averaged teacher score alone.
Instead, it first constructs the pair support from teacher-level score gaps, assigns the consensus direction, and then uses consensus confidence to down-weight conflicting comparisons.
CoRank contributes a multi-teacher procedure for constructing and weighting pairwise supervision before applying the pairwise logistic objective.
This trains the student network to act as a neural anomaly-scoring function whose scalar outputs reflect pairwise anomaly preferences supported by teacher consensus.

\vspace{-0.5em}
\subsection{Training and Inference}

Let $\mathcal{W}_{\mathrm{train}}=\{\mathbf{W}_t:t=L,\dots,T\}$ denote the set of training windows.
During training, each mini-batch $\mathcal{B}\subset\mathcal{W}_{\mathrm{train}}$ is paired with the normalized teacher scores at the corresponding endpoints, and AnoMix optimizes
\begin{equation}
    \theta^{\star}
    =
    \arg\min_{\theta}
    \mathbb{E}_{\mathcal{B}\sim \mathcal{W}_{\mathrm{train}}}
    \left[
    \mathcal{L}_{\mathrm{CoRank}}
    \left(
    g_{\theta};
    \mathcal{B},
    \{\tilde{z}^{(m)}\}_{m=1}^{M}
    \right)
    \right].
\end{equation}
At inference time, all teachers are discarded.
For each test window, the anomaly score is simply
\begin{equation}
    s_t
    =
    g_{\theta^{\star}}(\mathbf{W}_t),
    \qquad
    t=L,\dots,T_{\mathrm{test}}.
\end{equation}
Thus, pairwise training still yields a standard pointwise anomaly score at inference time.

\section{Experiments}
\label{sec:experiments}

\subsection{Setup}

\paragraph{Datasets.}
We evaluate on seven real-world multivariate TSAD benchmark datasets, including PSM~\citep{PSM-2021}, SWaT~\citep{SWaT-2016}, Exathlon~\citep{Exathlon-2021}, SMD~\citep{OmniAnomaly-2019}, SMAP~\citep{SMAP_MSL-2018}, Daphnet~\citep{Daphnet-2010}, and SWAN~\citep{SWAN-2020}.
Dataset descriptions and statistics are provided in Appendix~\ref{app:datasets}.
\vspace{-0.5em}
\paragraph{Metrics.}
We report six threshold-independent metrics: VUS-ROC and VUS-PR~\citep{VUS-2022}, AUC-ROC~\citep{ROC-2006}, AUC-PR~\citep{PR_ROC-2006}, and R-AUC-ROC and R-AUC-PR~\citep{PrecRec_for_TS-2018}.
The main experiments focus on VUS-ROC (V-ROC) and VUS-PR (V-PR).
Formal definitions of the reported metrics are provided in Appendix~\ref{app:metrics}.
\vspace{-0.5em}
\paragraph{Baselines.}
Table~\ref{tab:main} compares AnoMix against ten representative individual TSAD detectors.
These baselines include OmniAnomaly~\citep{OmniAnomaly-2019}, USAD~\citep{USAD-2020}, Anomaly Transformer~\citep{AnomalyTransformer-2021}, TranAD~\citep{TranAD-2022}, TimesNet~\citep{TimesNet-2022}, SARAD~\citep{SARAD-2024}, MOMENT~\citep{MOMENT-2024}, CATCH~\citep{CATCH-2025}, CrossAD~\citep{CrossAD-2025}, and OracleAD~\citep{OracleAD-2025}, covering the main TSAD paradigms considered in recent benchmarks.
Table~\ref{tab:ensemble} compares AnoMix against five score-ensemble baselines built from the same three teachers and the same normalized teacher scores.
We consider Mean, Max, Median, Rank Fusion, and Reciprocal Rank Fusion (RRF)~\citep{RRF-2009}.
For the zero-shot transfer setting in Table~\ref{tab:zero_shot}, we compare AnoMix-ZS against six pretrained time-series foundation models: MOMENT~\citep{MOMENT-2024}, TimesFM~\citep{TimesFM-2024}, Chronos~\citep{Chronos-2024}, MOIRAI~\citep{MOIRAI-2024}, Lag-Llama~\citep{LagLlama-2024}, and DADA~\citep{DADA-2025}.
For most deep TSAD and zero-shot baselines, we use implementations from the TSB-AD repository~\citep{Elephant_in_the_Room-2024}.
Baselines not available in TSB-AD are evaluated using their official repositories.
Baseline descriptions and the training and evaluation protocol are provided in Appendix~\ref{app:baselines}.

\vspace{-0.5em}
\paragraph{Teacher Set.}
By default, AnoMix uses a fixed three-teacher setting consisting of SARAD, CrossAD, and OracleAD.
This teacher set is formed by taking the three most recent state-of-the-art TSAD detectors in our teacher pool, which provides heterogeneous anomaly supervision while keeping the training-time teacher pool compact.
For analysis convenience and to avoid post-hoc teacher selection, this set was fixed before evaluating AnoMix on the benchmark datasets.
For each teacher, one fixed checkpoint supplies cached training scores shared across the five student seeds; Table~\ref{tab:main} separately reports each teacher's five-seed mean performance.
Alternative teacher-set choices, including leave-one-out (LOO) rankings and All-10, are analyzed in Section~\ref{sec:teacher_ablation}.
The LOO selection protocol and the full teacher ranking for each target dataset are detailed in Appendix~\ref{app:teacher_selection}.

\vspace{-0.5em}
\paragraph{Implementation Details.}
Retrained models use five random seeds under fixed splits; the zero-shot foundation baselines use fixed released checkpoints.
AnoMix uses a single-head TCN student with hidden dimension $64$, six causal blocks, kernel size $3$, dropout $0.1$, and fewer than 80k parameters on the benchmark datasets.
The student receptive field is 127 time steps, and most datasets use window size 127 to fully use this temporal context.
We train the student for 10 epochs using AdamW with cosine annealing and a score-gap threshold $\delta=0.1$ for valid-pair selection.
Student training uses only the provided training split and no anomaly labels.
Full implementation details are provided in Appendix~\ref{app:implementation_details}.

\vspace{-0.5em}
\subsection{Main Results}

\begin{table*}[t]
\centering
\caption{Main results on the seven benchmark datasets (5-seed mean, \%). Avg Rank is the per-row average rank across the 14 dataset-metric values, where lower is better. \textbf{Boldface} indicates the best result in each column, and \underline{underlined} values indicate the second-best result.}
\label{tab:main}
\small
\resizebox{\textwidth}{!}{
\setlength{\tabcolsep}{3pt}
\begin{tabular}{c|cc|cc|cc|cc|cc|cc|cc|c}
\toprule
\multirow{2}{*}{\textbf{Method}} & \multicolumn{2}{c|}{\textbf{PSM}} & \multicolumn{2}{c|}{\textbf{SWaT}} & \multicolumn{2}{c|}{\textbf{Exathlon}} & \multicolumn{2}{c|}{\textbf{SMD}} & \multicolumn{2}{c|}{\textbf{SMAP}} & \multicolumn{2}{c|}{\textbf{Daphnet}} & \multicolumn{2}{c|}{\textbf{SWAN}} & \multirow{2}{*}{\textbf{Avg Rank}} \\
\cmidrule(lr){2-3}\cmidrule(lr){4-5}\cmidrule(lr){6-7}\cmidrule(lr){8-9}\cmidrule(lr){10-11}\cmidrule(lr){12-13}\cmidrule(lr){14-15}
& \textbf{V-ROC} & \textbf{V-PR} & \textbf{V-ROC} & \textbf{V-PR} & \textbf{V-ROC} & \textbf{V-PR} & \textbf{V-ROC} & \textbf{V-PR} & \textbf{V-ROC} & \textbf{V-PR} & \textbf{V-ROC} & \textbf{V-PR} & \textbf{V-ROC} & \textbf{V-PR} & \\
\midrule
Omni~\citep{OmniAnomaly-2019} & 51.59 & 22.52 & 59.42 & 27.32 & \underline{99.98} & \underline{99.75} & 95.40 & 69.98 & 49.93 & 27.68 & 78.60 & 18.38 & 92.06 & 91.16 & 6.18 \\
USAD~\citep{USAD-2020} & 52.15 & 22.73 & 68.92 & 33.39 & 99.97 & 99.63 & 95.98 & 71.97 & 49.80 & 27.58 & 76.95 & 17.15 & 89.80 & 88.60 & 6.36 \\
A.T~\citep{AnomalyTransformer-2021} & 53.76 & 23.13 & 56.63 & 23.55 & 37.70 & 24.88 & 44.03 & 11.80 & 40.56 & 22.30 & 46.28 & 10.33 & 86.33 & 81.07 & 10.29 \\
TranAD~\citep{TranAD-2022} & 55.90 & 22.25 & \underline{77.06} & \underline{37.86} & 99.50 & 93.88 & 86.51 & 52.05 & 71.96 & 37.06 & 76.71 & 17.98 & 92.02 & 86.48 & 6.93 \\
TimesNet~\citep{TimesNet-2022} & 57.91 & 24.32 & 59.29 & 24.35 & 98.61 & 85.33 & 93.96 & 53.06 & 59.22 & 23.90 & 90.25 & 33.97 & \textbf{93.87} & 88.77 & 6.71 \\
SARAD~\citep{SARAD-2024} & 58.10 & 26.78 & 68.29 & 29.82 & 99.86 & 98.27 & 95.40 & 62.65 & 75.27 & 37.33 & 89.96 & 38.72 & 92.11 & 90.20 & \underline{4.25} \\
MOMENT~\citep{MOMENT-2024} & 42.39 & 16.59 & 54.95 & 20.21 & 99.66 & 97.17 & 57.27 & 10.19 & 54.72 & 27.85 & \underline{91.71} & 35.10 & 92.07 & 86.25 & 8.18 \\
\midrule
CATCH~\citep{CATCH-2025} & 53.99 & 23.42 & 58.82 & 24.61 & 99.83 & 97.79 & 96.45 & 67.33 & 49.69 & 27.45 & 91.37 & \underline{38.86} & 92.16 & 89.66 & 5.86 \\
CrossAD~\citep{CrossAD-2025} & 59.02 & 27.66 & 60.79 & 26.33 & 99.41 & 92.47 & \underline{96.49} & \underline{72.52} & 52.95 & 35.62 & 89.77 & 36.31 & 92.91 & \underline{91.66} & 4.79 \\
OracleAD~\citep{OracleAD-2025} & \underline{66.33} & \underline{28.68} & 59.87 & 26.44 & 99.66 & 97.18 & 85.71 & 52.69 & \underline{79.88} & \underline{40.18} & 88.46 & 34.60 & 92.44 & 86.30 & 5.39 \\
\midrule
\textbf{AnoMix} & \textbf{70.24} & \textbf{44.03} & \textbf{78.56} & \textbf{42.40} & \textbf{100.00} & \textbf{99.97} & \textbf{98.87} & \textbf{89.37} & \textbf{98.38} & \textbf{97.65} & \textbf{94.13} & \textbf{52.37} & \underline{93.75} & \textbf{91.76} & \textbf{1.07} \\
\bottomrule
\end{tabular}}
\end{table*}

\begin{table*}[t]
\centering
\caption{Score-ensemble comparison on the seven benchmark datasets (5-seed mean, \%). Avg Rank is the per-row average rank across the 14 dataset-metric values, where lower is better. \textbf{Boldface} indicates the best result in each column, and \underline{underlined} values indicate the second-best result.}
\label{tab:ensemble}
\small
\resizebox{\textwidth}{!}{
\setlength{\tabcolsep}{3pt}
\begin{tabular}{c|cc|cc|cc|cc|cc|cc|cc|c}
\toprule
\multirow{2}{*}{\textbf{Method}} & \multicolumn{2}{c|}{\textbf{PSM}} & \multicolumn{2}{c|}{\textbf{SWaT}} & \multicolumn{2}{c|}{\textbf{Exathlon}} & \multicolumn{2}{c|}{\textbf{SMD}} & \multicolumn{2}{c|}{\textbf{SMAP}} & \multicolumn{2}{c|}{\textbf{Daphnet}} & \multicolumn{2}{c|}{\textbf{SWAN}} & \multirow{2}{*}{\textbf{Avg Rank}} \\
\cmidrule(lr){2-3}\cmidrule(lr){4-5}\cmidrule(lr){6-7}\cmidrule(lr){8-9}\cmidrule(lr){10-11}\cmidrule(lr){12-13}\cmidrule(lr){14-15}
& \textbf{V-ROC} & \textbf{V-PR} & \textbf{V-ROC} & \textbf{V-PR} & \textbf{V-ROC} & \textbf{V-PR} & \textbf{V-ROC} & \textbf{V-PR} & \textbf{V-ROC} & \textbf{V-PR} & \textbf{V-ROC} & \textbf{V-PR} & \textbf{V-ROC} & \textbf{V-PR} & \\
\midrule
Mean       & 67.00 & 31.23 & 63.80 & 27.07 & 99.65 & 95.82 & 97.55 & 74.23 & 53.50 & 34.87 & 91.64 & 38.10 & 93.96 & 91.54 & 4.00 \\
Max        & 66.81 & 31.20 & 59.77 & 25.78 & 99.45 & 93.49 & 96.46 & 68.26 & 53.02 & 35.59 & 89.35 & 31.59 & \underline{93.98} & 91.30 & 5.21 \\
Median     & 61.93 & 27.28 & \underline{65.66} & 28.21 & 99.83 & 97.78 & 96.50 & 72.13 & 66.88 & 32.59 & 90.24 & 38.85 & 92.81 & 89.47 & 4.57 \\
Rank Fusion & 64.90 & 27.81 & 65.08 & \underline{28.57} & \underline{99.97} & \underline{99.74} & 97.72 & \underline{83.70} & 70.26 & 37.86 & \underline{92.71} & \underline{46.11} & 93.88 & \underline{91.69} & \underline{2.93} \\
RRF~\citep{RRF-2009} & \underline{67.37} & \underline{31.69} & 65.28 & 28.55 & 99.47 & 94.59 & \underline{97.73} & 79.13 & \underline{71.84} & \underline{39.62} & 91.32 & 36.18 & \textbf{94.24} & 91.59 & 3.00 \\
\midrule
\textbf{AnoMix} & \textbf{70.24} & \textbf{44.03} & \textbf{78.56} & \textbf{42.40} & \textbf{100.00} & \textbf{99.97} & \textbf{98.87} & \textbf{89.37} & \textbf{98.38} & \textbf{97.65} & \textbf{94.13} & \textbf{52.37} & 93.75 & \textbf{91.76} & \textbf{1.29} \\
\bottomrule
\end{tabular}}
\vspace{-0.5em}
\end{table*}

\paragraph{Benchmark Results.}
Table~\ref{tab:main} shows that AnoMix achieves the strongest overall average rank across the benchmark.
More importantly, strong baselines are not stable across datasets.
The second-best results are distributed across reconstruction, forecasting, association-aware, deviation-based, and foundation-model baselines, indicating that each detector family captures only part of the anomaly structure present in the benchmark.
AnoMix turns this dataset-dependent complementarity into a single student, rather than committing to one detector family at deployment.

The magnitude of improvement is also non-uniform across datasets and metrics.
AnoMix yields its largest gains on PSM, SMD, and especially SMAP, where the best individual detectors still leave substantial room for improvement, whereas the gains on Exathlon are small because several methods already achieve near-ceiling performance.
Across datasets, the improvement over the strongest individual baseline is especially pronounced on V-PR and smaller on V-ROC.
This suggests that AnoMix does not merely improve coarse ranking quality, but more sharply concentrates high anomaly scores on true anomalous regions.
Full PR and ROC metric results are reported in Appendix~\ref{app:full_metrics}.

\paragraph{Comparison with Score-Ensemble Baselines.}
Table~\ref{tab:ensemble} tests whether AnoMix's gains can be explained by static score aggregation alone.
The comparison fixes the teacher pool and changes only how teacher information is combined at inference time.
Among the fusion rules, RRF is the strongest overall, but AnoMix achieves a substantially better average rank.
The gaps are especially large on PSM, SWaT, SMD, and SMAP.
If the gains came merely from merging teacher outputs at test time, stronger fusion rules such as rank fusion or RRF would be expected to close most of the gap.
Instead, the gap persists even against rank-based fusion, indicating that AnoMix benefits from learning an input-dependent student scoring function rather than from static score aggregation alone.
Teacher scores are used as cross-teacher comparative supervision during training, allowing the student to learn anomaly preferences across teachers while deploying as a single model.

\subsection{Ablation Studies}
\label{sec:ablation}

\begin{figure}[t]
    \centering
    \includegraphics[width=0.98\textwidth]{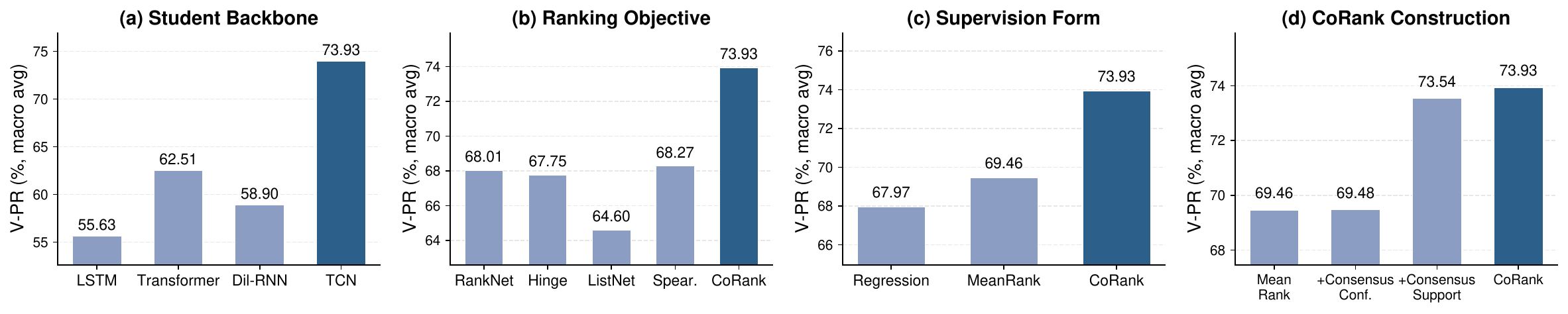}
    \caption{Ablation study reporting macro-average V-PR (\%) over the seven benchmark datasets.
      (a) Student backbones: LSTM, Transformer, Dilated-RNN, and TCN.
      (b) Ranking losses: RankNet, Hinge, ListNet, and Spearman on an averaged teacher-score target, and CoRank.
      (c) Supervision forms: score regression on the averaged teacher-score target, MeanRank, and CoRank.
      (d) CoRank decomposition: MeanRank, consensus confidence, consensus support, and full CoRank.}
    \label{fig:ablation}
    \vspace{-0.5em}
\end{figure}

\paragraph{Backbone Choice.}
To assess backbone dependence, Figure~\ref{fig:ablation}(a) compares four student backbones under the same CoRank objective.
Each backbone is used within the same endpoint-scoring template, producing an endpoint representation that is mapped to a scalar anomaly score.
The LSTM, Transformer, and Dilated-RNN variants use their best learning rate on each dataset.
TCN achieves the highest macro V-PR (73.93\%), outperforming Transformer (62.51\%), Dilated-RNN (58.90\%), and LSTM (55.63\%) by $11$--$18$\%pt.
With a comparable number of parameters, the result supports choosing a TCN as an efficient lightweight student for distilling pairwise anomaly preferences, without introducing a new architecture design.
\vspace{-0.5em}
\paragraph{Training Objective.}
To test whether standard ranking losses suffice, Figure~\ref{fig:ablation}(b) compares CoRank with four such losses using the same student.
The baseline losses rank pairs according to the averaged normalized teacher score, but they do not use CoRank's consensus support or consensus confidence.
CoRank achieves the highest macro V-PR (73.93\%), exceeding RankNet~\citep{RankNet-2005} (68.01\%), Hinge~\citep{LearningToRank-2009} (67.75\%), ListNet~\citep{ListNet-2007} (64.60\%), and Spearman~\citep{Blondel-2020} (68.27\%).
Because the student architecture is held fixed, the 5--10\%pt gap is not attributable to model capacity or to the use of ranking supervision alone.
It reflects the difference between applying standard ranking surrogates to a single fused target and constructing the pairwise training signal from teacher-level score gaps.
In this setting, CoRank's advantage comes from how pair support and confidence are defined before the logistic ranking surrogate is applied.
\vspace{-0.5em}
\paragraph{Score Supervision Form.}
To assess the benefit of pairwise supervision, Figure~\ref{fig:ablation}(c) contrasts direct score regression, fused-target ranking, and CoRank using the same single-head student.
In the score-regression baseline, the student directly fits the averaged normalized teacher score using a pointwise MSE loss.
MeanRank denotes the fused-target ranking baseline that first averages normalized teacher scores and then forms ranking pairs from gaps in the averaged score.
This baseline reaches 67.97\%, MeanRank reaches 69.46\%, and CoRank reaches 73.93\%.
These results support two conclusions.
First, relative supervision is more effective than direct pointwise score matching.
Second, ranking the averaged teacher score is not sufficient to recover the full benefit of multi-teacher distillation.
This comparison shows that the benefit comes not only from ranking, but also from constructing pairwise supervision from teacher-level score gaps before averaging.
Figure~\ref{fig:ablation}(d) further decomposes this teacher-level construction.
Together with Figure~\ref{fig:ablation}(b), these results suggest that AnoMix benefits specifically from consensus-based multi-teacher distillation rather than from ranking alone.
\vspace{-0.5em}
\paragraph{CoRank Component Decomposition.}
To identify which components drive CoRank's gains, Figure~\ref{fig:ablation}(d) separates the effects of consensus support and consensus confidence while using the same pairwise logistic loss across all variants.
MeanRank reaches 69.46\% macro V-PR, and adding consensus confidence under the same support from averaged-score gaps changes this only slightly to 69.48\%.
Using CoRank's consensus support increases macro V-PR to 73.54\%, and adding consensus confidence on top yields full CoRank at 73.93\%.
This indicates that CoRank's main gain comes from constructing the pairwise training set from teacher-level score gaps before averaging.
Averaging teachers before support construction can weaken comparisons where one detector gives a clear score gap but other detectors give small or neutral gaps, whereas consensus support preserves these comparisons for the student.
Consensus confidence then provides a smaller additional gain by reducing the influence of retained pairs with stronger teacher conflict.
Additional ablation studies are provided in Appendix~\ref{app:additional_ablations}.

\subsection{Distillation Analysis}
\label{sec:distillation_evidence}
\label{sec:teacher_ablation}
\begin{table*}[t]
\centering
\caption{Teacher-set variants on the seven benchmark datasets (5-seed mean, \%; standard deviations in Appendix~\ref{app:significance}). LOO-$k$ uses the top-$k$ teachers from the target-excluded teacher ranking, and Sel-3 is the fixed three-teacher setting used in Table~\ref{tab:main}. Avg Rank is computed over the 14 reported values, where lower is better. \textbf{Boldface} indicates the best result in each column, and \underline{underlined} values indicate the second-best result.}
\label{tab:teacher_sets}
\footnotesize
\resizebox{\textwidth}{!}{
\setlength{\tabcolsep}{3pt}
\begin{tabular}{l|cc|cc|cc|cc|cc|cc|cc|c}
\toprule
\multirow{2}{*}{\textbf{Method}} & \multicolumn{2}{c|}{\textbf{PSM}} & \multicolumn{2}{c|}{\textbf{SWaT}} & \multicolumn{2}{c|}{\textbf{Exathlon}} & \multicolumn{2}{c|}{\textbf{SMD}} & \multicolumn{2}{c|}{\textbf{SMAP}} & \multicolumn{2}{c|}{\textbf{Daphnet}} & \multicolumn{2}{c|}{\textbf{SWAN}} & \multirow{2}{*}{\textbf{Avg Rank}} \\
\cmidrule(lr){2-3} \cmidrule(lr){4-5} \cmidrule(lr){6-7} \cmidrule(lr){8-9} \cmidrule(lr){10-11} \cmidrule(lr){12-13} \cmidrule(lr){14-15}
& \textbf{V-ROC} & \textbf{V-PR} & \textbf{V-ROC} & \textbf{V-PR} & \textbf{V-ROC} & \textbf{V-PR} & \textbf{V-ROC} & \textbf{V-PR} & \textbf{V-ROC} & \textbf{V-PR} & \textbf{V-ROC} & \textbf{V-PR} & \textbf{V-ROC} & \textbf{V-PR} & \\
\midrule
LOO-1  & 48.10 & 22.32 & 43.48 & 25.09 & 80.05 & 81.58 & 59.08 & 48.49 & 56.59 & 56.58 & 75.65 & 30.98 & 90.81 & 84.37 & 7.00 \\
LOO-2  & \underline{69.91} & 42.22 & 73.28 & 38.72 & 99.99 & 99.96 & \underline{98.92} & \textbf{89.48} & 98.37 & 97.65 & 94.01 & 51.50 & 92.45 & 90.77 & 4.61 \\
LOO-3  & 69.04 & \underline{42.64} & 76.82 & 43.37 & 99.99 & 99.96 & 98.87 & 89.37 & 98.40 & 97.72 & 94.12 & 51.96 & \textbf{93.75} & \textbf{91.76} & 3.61 \\
LOO-4  & 67.55 & 35.17 & \textbf{80.98} & \underline{44.59} & 99.99 & 99.96 & 98.90 & 89.37 & 98.41 & 97.75 & 94.10 & 51.94 & 93.60 & 91.52 & 3.68 \\
LOO-5  & 67.64 & 34.40 & 75.49 & 41.28 & 99.99 & 99.96 & 98.90 & 89.39 & \underline{98.42} & \underline{97.77} & \underline{94.15} & 52.25 & 93.43 & 91.28 & 3.68 \\
Sel-3  & \textbf{70.24} & \textbf{44.03} & 78.56 & 42.40 & \underline{100.00} & \underline{99.97} & 98.87 & 89.37 & 98.38 & 97.65 & 94.13 & \textbf{52.37} & \textbf{93.75} & \textbf{91.76} & \underline{2.86} \\
All-10 & 64.45 & 31.19 & \underline{80.47} & \textbf{45.23} & \underline{100.00} & \underline{99.97} & \textbf{98.93} & \underline{89.41} & \textbf{98.43} & \textbf{97.78} & \textbf{94.17} & \underline{52.34} & 93.23 & 91.06 & \textbf{2.57} \\
\bottomrule
\end{tabular}}
\end{table*}
\begin{figure}[t]
    \centering
    \includegraphics[width=0.95\textwidth]{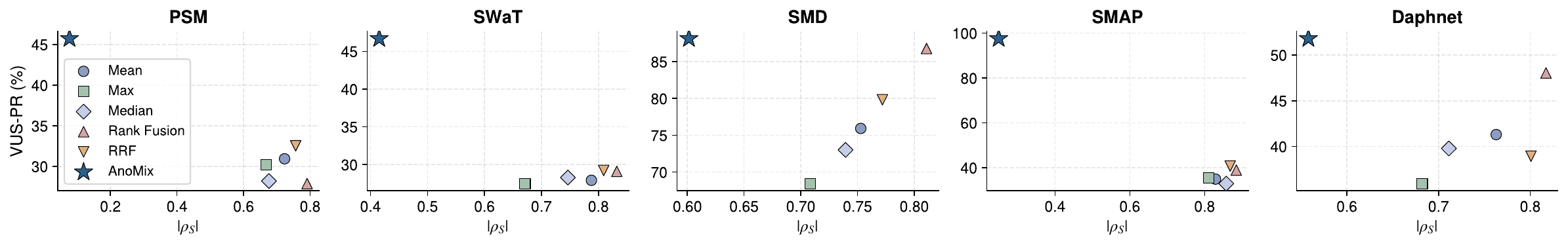}
    \caption{Mean absolute Spearman correlation with teacher score sequences ($|\rho_S|$) versus V-PR on five non-saturated datasets. A higher $|\rho_S|$ indicates stronger alignment with teacher score sequences, while a lower $|\rho_S|$ indicates weaker alignment. The blue star denotes AnoMix, and the remaining markers denote the score-ensemble baselines (Mean, Max, Median, Rank Fusion, RRF).}
    \vspace{-0.5em}
    \label{fig:ranking_effect}
\end{figure}
\paragraph{Teacher-Set Variants.}
We test AnoMix's sensitivity to teacher identity and pool size.
For each target dataset, LOO-$k$ selects the top $k$ teachers by average V-PR on the other six datasets; Sel-3 is the fixed three-teacher set chosen before evaluation, and All-10 uses the full teacher pool.
Table~\ref{tab:teacher_sets} shows a clear separation between single-teacher and multi-teacher settings.
LOO-1 is substantially weaker than all evaluated multi-teacher variants, suggesting that the teacher signals are complementary enough for AnoMix to benefit from combining them rather than relying on a single teacher.
This is consistent with the role of CoRank, which aggregates teacher-supported score gaps instead of matching any single teacher score directly.

The same table also shows that AnoMix is not highly sensitive to teacher-set selection.
All-10 achieves the best overall average rank, while the fixed three-teacher benchmark setting (Sel-3) remains close to the stronger LOO variants.
However, All-10 also degrades PSM substantially, suggesting that broader teacher pools can add complementary training signal while also introducing harmful or redundant comparisons.
The similar performance of LOO-3, LOO-4, LOO-5, and Sel-3 suggests that a compact heterogeneous pool captures most of the gain, with LOO selection offering a practical alternative.
Appendix~\ref{app:teacher_selection} further tests mixed-quality, weak, and correlated teacher pools.

\vspace{-0.5em}

\paragraph{Score Alignment.}
Figure~\ref{fig:ranking_effect} tests whether AnoMix is merely approximating a static ensemble.
Across the non-saturated datasets, AnoMix attains higher V-PR than all score-ensemble baselines while exhibiting lower average Spearman correlation with the teacher score sequences.
If the student were simply learning a smoothed version of one teacher or of a fused teacher score, stronger performance would be expected to come with stronger teacher-score alignment.
Instead, the opposite pattern is observed: AnoMix is more accurate while being less correlated with teacher outputs.
This indicates that the student is not passively inheriting teacher scores, but integrating teacher information into a more effective neural anomaly-scoring function.
Taken together with Table~\ref{tab:teacher_sets}, this supports the intended role of CoRank as a distillation objective rather than a surrogate ensemble rule.
Score visualizations and teacher-alignment diagnostics are provided in Appendices~\ref{app:score_visualization} and~\ref{app:teacher_alignment}.
\vspace{-0.5em}

\subsection{Zero-Shot Cross-Dataset Detection}
\label{sec:zero_shot}
\begin{table*}[t]
\centering
\caption{Zero-shot transfer results (\%). AnoMix-ZS reports five-seed means, with standard deviations in Appendix~\ref{app:significance}; foundation baselines use fixed released checkpoints. AnoMix-ZS is trained on the other six datasets and evaluated on the held-out target. \textbf{Boldface} indicates the best result in each column, and \underline{underlined} values indicate the second-best result.}
\label{tab:zero_shot}
\footnotesize
\resizebox{\textwidth}{!}{
\setlength{\tabcolsep}{2.5pt}
\begin{tabular}{c|c|cc|cc|cc|cc|cc|cc|cc|c}
\toprule
\multirow{2}{*}{\textbf{Method}} & \multirow{2}{*}{\textbf{Params}} & \multicolumn{2}{c|}{\textbf{PSM}} & \multicolumn{2}{c|}{\textbf{SWaT}} & \multicolumn{2}{c|}{\textbf{Exathlon}} & \multicolumn{2}{c|}{\textbf{SMD}} & \multicolumn{2}{c|}{\textbf{SMAP}} & \multicolumn{2}{c|}{\textbf{Daphnet}} & \multicolumn{2}{c|}{\textbf{SWAN}} & \multirow{2}{*}{\textbf{Avg Rank}} \\
\cmidrule(lr){3-4}\cmidrule(lr){5-6}\cmidrule(lr){7-8}\cmidrule(lr){9-10}\cmidrule(lr){11-12}\cmidrule(lr){13-14}\cmidrule(lr){15-16}
 & & \textbf{V-ROC} & \textbf{V-PR} & \textbf{V-ROC} & \textbf{V-PR} & \textbf{V-ROC} & \textbf{V-PR} & \textbf{V-ROC} & \textbf{V-PR} & \textbf{V-ROC} & \textbf{V-PR} & \textbf{V-ROC} & \textbf{V-PR} & \textbf{V-ROC} & \textbf{V-PR} & \\
\midrule
MOMENT~\citep{MOMENT-2024}   & 110M & 41.87 & 16.66 & 54.62 & 20.17 & 99.66 & 97.18 & 58.22 & 10.38 & 56.03 & 28.31 & 91.49 & 34.49 & 92.06 & 86.29 & 5.57 \\
TimesFM~\citep{TimesFM-2024}  & 200M & 53.99 & 20.14 & 49.45 & 19.22 & 99.19 & 96.64 & 59.04 & 16.19 & 94.65 & 88.93 & 89.97 & 34.79 & \textbf{93.79} & 87.66 & 4.50 \\
Chronos~\citep{Chronos-2024}  & 710M & 52.46 & 18.68 & \underline{63.11} & 23.33 & 87.54 & 61.15 & 89.06 & 34.67 & 70.54 & 30.97 & 90.53 & 33.50 & 93.02 & 88.11 & 4.71 \\
MOIRAI~\citep{MOIRAI-2024}   & 14M  & 54.02 & 19.47 & 55.17 & 20.49 & 91.52 & 66.86 & 60.99 & 10.37 & 57.98 & 21.12 & 91.07 & 34.11 & 92.64 & 83.72 & 5.43 \\
Lag-Llama~\citep{LagLlama-2024} & 10M & 54.63 & 20.45 & 47.96 & 18.56 & 89.08 & 58.11 & 83.10 & 33.47 & 80.29 & 39.28 & 92.04 & 38.24 & 93.11 & 87.86 & 4.43 \\
DADA~\citep{DADA-2025}    & 1.84M & \underline{57.80} & \underline{23.08} & 60.82 & \underline{25.40} & \underline{99.80} & \textbf{98.72} & \textbf{98.62} & \textbf{87.90} & \textbf{98.67} & \textbf{95.61} & \underline{92.53} & \underline{40.29} & \underline{93.57} & \underline{88.84} & \underline{1.71} \\
\midrule
\textbf{AnoMix-ZS} & \textbf{83k} & \textbf{58.71} & \textbf{25.44} & \textbf{78.97} & \textbf{39.85} & \textbf{99.88} & \underline{98.19} & \underline{98.45} & \underline{86.12} & \underline{98.19} & \underline{93.88} & \textbf{93.53} & \textbf{47.73} & 92.91 & \textbf{90.28} & \textbf{1.64} \\
\bottomrule
\end{tabular}}
\vspace{-0.5em}
\end{table*}
AnoMix's training procedure suggests a natural zero-shot extension because the student can be distilled from teacher-derived ranking signals on source datasets and then applied to an unseen target dataset.
We evaluate this extension under a stricter cross-dataset setting that removes target-dataset supervision entirely.
For this evaluation, we use AnoMix-ZS, a compact channel-independent student that can operate on varying input dimensionality.
The architecture and protocol are detailed in Appendix~\ref{app:student_architectures}.
AnoMix-ZS is trained on six datasets with the same ranking-based distillation objective and then evaluated on the held-out target, without observing the target dataset or teacher scores during training.
The six foundation baselines are converted into anomaly scores following their evaluation wrappers: MOMENT uses reconstruction error, TimesFM, Chronos, MOIRAI, and Lag-Llama use one-step forecasting error, and DADA uses mask-reconstruction variance.

Table~\ref{tab:zero_shot} shows that ranking-based distillation transfers beyond the training benchmarks and remains highly competitive even in the strict zero-shot setting.
Among the evaluated models, AnoMix-ZS achieves the best average rank ($1.64$) with $83$k parameters, whereas the pretrained foundation baselines range from $1.84$M to $710$M parameters.
This comparison is informative for two reasons.
First, AnoMix-ZS is not merely memorizing source-dataset behavior, but learning anomaly-ranking information that generalizes to unseen datasets.
Second, the comparison shows that cross-dataset TSAD-oriented distillation can match a much larger TSAD-specific foundation model ($83$k vs $1.84$M parameters for DADA), and substantially outperforms general-purpose time-series foundation models that range from $10$M to $710$M parameters.
Taken together, these results suggest that TSAD-oriented cross-dataset distillation, as in AnoMix, offers a promising direction for building TSAD-specific foundation models rather than relying solely on general-purpose time-series foundation models.

\section{Conclusion}

We studied whether unsupervised time series anomaly detection can be effectively addressed through knowledge distillation despite the absence of labels and the heterogeneity of teacher anomaly scores.
Our experimental results suggest that it can.
We introduced AnoMix, a multi-teacher distillation framework that learns a single lightweight student by transferring pairwise anomaly preferences from multiple heterogeneous TSAD teachers, rather than directly imitating raw teacher scores.

Across benchmarks, the empirical evidence consistently shows that the benefit of multi-teacher TSAD comes from distilling complementary teacher scoring principles into a new student predictor, not merely from combining teacher scores.
AnoMix outperforms individual detectors and score-ensemble baselines, while teacher-set and correlation analyses further indicate that it learns beyond static ensemble behavior and remains effective under broader teacher pools.
The zero-shot cross-dataset results further show that AnoMix-ZS remains effective on unseen datasets and can compete with much larger pretrained time-series foundation models.

Although teachers use complex, detector-specific models, their scoring principles are ultimately expressed through scalar anomaly scores.
Our results suggest that learning pairwise preferences from these scores allows a compact student to integrate complementary scoring principles without reproducing the teachers' internal models.
A more detailed discussion of limitations and future directions is provided in Appendix~\ref{app:limitations}.

\bibliographystyle{unsrtnat}
\bibliography{reference}

\newpage
\appendix
\setcounter{table}{0}
\setcounter{figure}{0}
\makeatletter
\renewcommand{\thetable}{S\@arabic\c@table}
\renewcommand{\thefigure}{S\@arabic\c@figure}
\renewcommand{\theHtable}{S\@arabic\c@table}
\renewcommand{\theHfigure}{S\@arabic\c@figure}
\makeatother

\section{Rationale for Multi-Teacher Ranking Distillation}
\label{app:theory}
\label{app:rationale}
\label{app:ranking-distillation-theory}

AnoMix trains a neural anomaly-scoring function $g_\theta:\mathcal X\rightarrow\mathbb R$.
For notational simplicity, we write $x_i$ for the training window sample associated with an endpoint.
Since the student outputs a scalar score, learning this function does not require a pointwise anomaly-score target.
It requires supervision that shapes the ordering induced by $g_\theta$.
In supervised anomaly detection, one could train the student against binary anomaly labels or labeled pairwise preferences.
In unsupervised TSAD, however, no such labels are available.
The only training signal comes from frozen teacher detectors.

A direct pointwise regression target is not identifiable in this setting.
Each teacher produces a scalar anomaly score, but the score is generated from a detector-specific anomaly-scoring principle, such as reconstruction error, forecasting residual, association discrepancy, or representation deviation.
Even after per-teacher normalization, these scores do not share a calibrated score scale.
Direct regression can therefore spend student capacity fitting teacher-specific score scales rather than learning pairwise anomaly preferences supported by multiple teachers.
Moreover, for any strictly increasing transformation $h_m$,
\begin{equation}
    s_m(x_i) > s_m(x_j)
    \quad \Longleftrightarrow \quad
    h_m(s_m(x_i)) > h_m(s_m(x_j)).
\end{equation}
The teacher-induced ordering is preserved, but the pointwise regression target is changed.
Thus, without anomaly labels, the score scale cannot be identified from data alone.
AnoMix therefore distills pairwise anomaly preferences from teacher score gaps rather than treating teacher scores as pointwise regression targets.

Let $\tilde z_i^{(m)}$ be the normalized score assigned by teacher $m$ to sample $x_i$, and define the teacher score-gap vector for a pair $(i,j)$ as
\begin{equation}
    \boldsymbol{\Delta}_{ij}
    =
    \left(
    \Delta_{ij}^{(1)},\ldots,\Delta_{ij}^{(M)}
    \right),
    \qquad
    \Delta_{ij}^{(m)}
    =
    \tilde z_i^{(m)}-\tilde z_j^{(m)}.
\end{equation}
Training a neural anomaly-scoring function from these teacher score gaps requires converting $\boldsymbol{\Delta}_{ij}$ into pairwise supervision.
We write this supervision as
\begin{equation}
    \Gamma(\boldsymbol{\Delta}_{ij})
    =
    (A_{ij},y_{ij},c_{ij}),
\end{equation}
where $A_{ij}\in\{0,1\}$ indicates whether the pair is informative enough to use, $y_{ij}\in\{-1,0,+1\}$ specifies the consensus pairwise direction, and $c_{ij}\in[0,1]$ controls the strength of the pair in training.
Given such supervision, a natural pairwise learning objective is
\begin{equation}
    \mathcal L(\theta)
    =
    \frac{1}{\sum_{i<j}A_{ij}}
    \sum_{i<j}
    A_{ij}c_{ij}
    \log
    \left(
    1+
    \exp
    \left(
    -y_{ij}
    \left(g_\theta(x_i)-g_\theta(x_j)\right)
    \right)
    \right).
\end{equation}
If no pair is retained in a mini-batch, the batch contributes zero loss.
This objective directly trains the student score margin $g_\theta(x_i)-g_\theta(x_j)$ rather than regressing to teacher-specific pointwise scores.

The label-free multi-teacher setting imposes three requirements on $\Gamma(\boldsymbol{\Delta}_{ij})$.
First, it should select informative pairs using teacher-level score gaps, because averaging teacher scores before pair selection can suppress a clear gap from one detector when other detectors give small or neutral gaps.
Second, it should produce a scalar-compatible consensus direction, because the student itself outputs a single scalar anomaly score and can only realize scalar score orderings.
Third, it should be conflict-aware, so that pairs with strong opposing teacher score gaps do not contribute as much as pairs with little directional conflict.

CoRank instantiates $\Gamma$ with the following choices.
The support indicator is
\begin{equation}
    A_{ij}
    =
    \mathbf{1}
    \left[
    \|\boldsymbol{\Delta}_{ij}\|_\infty>\delta
    \right]
    =
    \mathbf{1}
    \left[
    \max_m|\Delta_{ij}^{(m)}|>\delta
    \right].
\end{equation}
This retains a pair whenever at least one teacher observes a meaningful score gap.
The condition does not assume that the teacher producing the gap is correct.
It only marks the pair as meaningful for consensus supervision before teacher-level contrasts are suppressed by averaging.

The consensus direction is
\begin{equation}
    y_{ij}
    =
    \operatorname{sgn}
    \left(
    \mathbf{1}^\top\boldsymbol{\Delta}_{ij}
    \right)
    =
    \operatorname{sgn}
    \left(
    \sum_{m=1}^{M}\Delta_{ij}^{(m)}
    \right).
\end{equation}
This direction is equivalent to comparing the averaged normalized teacher score
\begin{equation}
    \operatorname{sgn}
    \left(
    \sum_{m=1}^{M}\Delta_{ij}^{(m)}
    \right)
    =
    \operatorname{sgn}
    \left(
    \bar z_i-\bar z_j
    \right),
    \qquad
    \bar z_i
    =
    \frac{1}{M}\sum_{m=1}^{M}\tilde z_i^{(m)}.
\end{equation}
This equivalence defines only the consensus direction after the support set has been constructed from teacher-level gaps.
Thus, CoRank does not ask the student to fit an arbitrary cyclic preference relation.
The pairwise direction is induced by a scalar teacher-consensus score and is compatible with the student's scalar output.

The confidence weight is
\begin{equation}
    c_{ij}
    =
    \frac{
    \left|
    \mathbf{1}^\top\boldsymbol{\Delta}_{ij}
    \right|
    }{
    \|\boldsymbol{\Delta}_{ij}\|_1
    }
    =
    \frac{
    \left|
    \sum_{m=1}^{M}\Delta_{ij}^{(m)}
    \right|
    }{
    \sum_{m=1}^{M}|\Delta_{ij}^{(m)}|
    }.
\end{equation}
By the triangle inequality, $0\le c_{ij}\le 1$ for every retained pair.
This weight measures directional conflict among teacher score gaps.
It is large when teacher score gaps provide little opposition to the consensus direction, and small when teacher score gaps strongly conflict.
Importantly, $c_{ij}$ should not be interpreted as an oracle estimate of teacher correctness.
It is a label-free conflict weight.
If the consensus direction is zero, the numerator is zero and the pair does not affect the update.

Substituting these choices into the general pairwise objective and writing $\mathcal P=\{(i,j):A_{ij}=1\}$ yields
\begin{equation}
    \mathcal L_{\mathrm{CoRank}}
    =
    \frac{1}{|\mathcal P|}
    \sum_{(i,j)\in\mathcal P}
    c_{ij}
    \log
    \left(
    1+
    \exp
    \left(
    -y_{ij}
    \left(g_\theta(x_i)-g_\theta(x_j)\right)
    \right)
    \right).
\end{equation}
Therefore, CoRank provides a concrete way to convert heterogeneous teacher score gaps into pairwise supervision for training a neural anomaly-scoring function.
Compared with pointwise regression, it avoids fitting a non-identifiable score scale.
Compared with single-teacher ranking, it uses comparative signals from multiple anomaly-scoring principles.
Compared with ranking on an averaged teacher score, it selects informative pairs from teacher-level score gaps before aggregation and then discounts directional teacher conflict.

\clearpage
\section{Student Architectures}
\label{app:student_architectures}

\paragraph{Endpoint Scoring.}
Both the main AnoMix student and AnoMix-ZS follow the same prediction template.
Given an input window ending at time $t$, the student first constructs an endpoint-centered representation $\mathbf{h}_t \in \mathbb{R}^{H}$ and then maps it to one scalar anomaly score:
\begin{equation}
    \mathbf{h}_t = \Psi(\mathbf{W}_t),
    \qquad
    \hat{z}_t = \mathbf{w}_o^{\top}\mathbf{h}_t + b_o.
\end{equation}
This template matches the supervision used by AnoMix, since teacher scores and CoRank pairs are defined for window endpoints.
The main student and the zero-shot student differ only in the representation map $\Psi(\cdot)$.

\paragraph{Fixed-Dimensional Encoder.}
For the main benchmark setting, all train and test windows within a dataset have the same number of variables $D$.
We therefore use a compact multivariate causal TCN encoder.
Given $\mathbf{W}_t = (\mathbf{x}_{t-L+1}, \dots, \mathbf{x}_t) \in \mathbb{R}^{L \times D}$, each timestep is projected into the hidden dimension:
\begin{equation}
    \mathbf{h}^{(0)}_i = \mathbf{P}^{\top}\mathbf{x}_i + \mathbf{b}_p \in \mathbb{R}^{H},
    \qquad i = t-L+1,\dots,t,
\end{equation}
which defines $\mathbf{H}^{(0)} = (\mathbf{h}^{(0)}_{t-L+1}, \dots, \mathbf{h}^{(0)}_t) \in \mathbb{R}^{L \times H}$.
The temporal encoder applies $N$ causal convolutional blocks:
\begin{equation}
    \mathbf{H}^{(\ell)} = \Phi_{\ell}\!\left(\mathbf{H}^{(\ell-1)}\right),
    \qquad \ell = 1,\dots,N.
\end{equation}
Here, $\mathbf{P} \in \mathbb{R}^{D \times H}$ and $\mathbf{b}_p \in \mathbb{R}^{H}$ define the input projection, and $\Phi_{\ell}$ denotes a causal temporal block.
The endpoint representation used by the shared scoring template is
\begin{equation}
    \mathbf{h}_t = \mathbf{H}^{(N)}_L.
\end{equation}
In the main paper, we use $N=6$ blocks with dilations $\{1,2,4,8,16,32\}$ and kernel size $k=3$.
Each block consists of a causal 1D convolution, residual connection, GELU activation, and dropout.
The resulting receptive field is 127 when $L=127$, so the endpoint representation can summarize the full input window before the final scalar prediction is made.
Under the benchmark settings used in the main paper, this student contains under 80k parameters.

\paragraph{Variable-Dimensional Encoder.}
In zero-shot transfer, the held-out dataset may have a different number of variables from the source datasets.
The main multivariate encoder is tied to a fixed input dimension $D$, because its first projection maps $\mathbb{R}^{D}$ to $\mathbb{R}^{H}$.
AnoMix-ZS keeps the same endpoint-scoring template but replaces this $D$-dependent input encoder with a channel-independent encoder.
It processes each variable with shared parameters, constructs one endpoint representation per variable, and aggregates those representations into the single $\mathbf{h}_t$ used by the score head.

Let $\mathbf{x}^{(d)} = (x_{t-L+1,d}, \dots, x_{t,d}) \in \mathbb{R}^{L \times 1}$ denote the univariate window for variable $d$.
All variables are reshaped into the batch dimension and passed through the same univariate projection:
\begin{equation}
    \mathbf{h}^{(0)}_{d,i} = \phi(x_{i,d}) \in \mathbb{R}^{H},
    \qquad d=1,\dots,D.
\end{equation}
A fixed sinusoidal positional encoding is added, followed by a single-layer Pre-LayerNorm Transformer and four causal TCN blocks with dilations $\{1,2,4,8\}$.
This produces a per-variable endpoint representation $\mathbf{g}_{t,d} \in \mathbb{R}^{H}$.
AnoMix-ZS then aggregates variables with learned attention weights:
\begin{equation}
    a_{t,d}
    =
    \frac{\exp(\mathbf{w}_a^{\top}\mathbf{g}_{t,d})}
    {\sum_{r=1}^{D}\exp(\mathbf{w}_a^{\top}\mathbf{g}_{t,r})},
    \qquad
    \mathbf{h}_t
    =
    \sum_{d=1}^{D} a_{t,d}\mathbf{g}_{t,d}.
\end{equation}
The same scalar score head then computes $\hat{z}_t = \mathbf{w}_o^{\top}\mathbf{h}_t + b_o$.
Thus, AnoMix-ZS changes only the encoder used to obtain the endpoint representation, while preserving the endpoint-centered scoring template and the CoRank objective.
With $H=64$, one Transformer layer, four TCN blocks, and one score head, AnoMix-ZS has 83k parameters independent of $D$.

\paragraph{Zero-Shot Training.}
AnoMix-ZS uses the same CoRank objective as the main method.
For each target dataset $\mathcal{D}^{\star}$, the student is trained on the concatenation of training windows from the remaining six benchmark datasets and evaluated on $\mathcal{D}^{\star}$ with no target-dataset fine-tuning, no target teacher scores, and no target training data.
Teacher scores on the training mixture are obtained from the same stored per-dataset teacher checkpoints used in the main experiments, and CoRank supervision is constructed only within each dataset's mini-batch so that teacher normalizations remain dataset-consistent.

\clearpage
\section{Pseudocode}
\label{app:pseudocode}

Algorithm~\ref{alg:anomix} summarizes the full AnoMix pipeline used in the main paper.
It separates offline teacher preprocessing, single-head student training with CoRank, and teacher-free inference.

\begin{algorithm}[h]
\caption{AnoMix Training and Inference}
\label{alg:anomix}
\footnotesize
\begin{algorithmic}[1]
\Statex \textbf{Input:} $\mathbf{X}_{\mathrm{train}} \in \mathbb{R}^{T \times D}$, $\mathbf{X}_{\mathrm{test}} \in \mathbb{R}^{T' \times D}$, teachers $\{f_m\}_{m=1}^{M}$, window length $L$, batch size $B$, epochs $E$, learning rate $\eta$, threshold $\delta$
\Statex \textbf{Output:} Student $g_\theta$ and anomaly scores $\{s_t\}_{t=L}^{T'}$
\Statex \textit{Stage 1. Offline teacher preprocessing}
\For{$m = 1,\dots,M$}
    \State Compute teacher scores $z^{(m)} \gets f_m(\mathbf{X}_{\mathrm{train}})$
    \State Min-max normalize $z^{(m)}$ on $\mathbf{X}_{\mathrm{train}}$ to obtain $\tilde z^{(m)}$
\EndFor
\Statex \textit{Stage 2. Single-head student training with CoRank}
\State $\mathcal{D} \gets \{(\mathbf{W}_t, \tilde{\mathbf{z}}_t)\}_{t=L}^{T}$, where $\tilde{\mathbf{z}}_t=(\tilde z_t^{(1)},\dots,\tilde z_t^{(M)})$
\State Initialize $g_\theta:\mathbb{R}^{L\times D}\to\mathbb{R}$ as the TCN student
\State Initialize AdamW with learning rate $\eta$ and cosine annealing for $E$ epochs
\For{$e = 1,\dots,E$}
    \For{mini-batch $\mathcal{B} = \{(\mathbf{W}_i,\tilde{\mathbf{z}}_i)\}_{i=1}^{B} \sim \mathcal{D}$}
        \State $\hat z_i \gets g_\theta(\mathbf{W}_i)$ for all $i \in [B]$
        \State Enumerate all unordered pairs $(i,j)$ with $i<j$
        \State $\boldsymbol{\Delta}_{ij} \gets \tilde{\mathbf{z}}_i-\tilde{\mathbf{z}}_j$ for each pair
        \State $\mathcal{P} \gets \{(i,j): i<j,\ \max_m |\Delta_{ij}^{(m)}|>\delta\}$
        \If{$\mathcal{P}=\emptyset$}
            \State \textbf{continue}
        \EndIf
        \For{each $(i,j)\in\mathcal{P}$}
            \State $y_{ij} \gets \operatorname{sgn}\left(\sum_{m=1}^{M}\Delta_{ij}^{(m)}\right)$
            \State $c_{ij} \gets \left|\sum_{m=1}^{M}\Delta_{ij}^{(m)}\right| / \sum_{m=1}^{M}|\Delta_{ij}^{(m)}|$
            \State $d_{ij} \gets \hat z_i-\hat z_j$
        \EndFor
        \State $\mathcal{L}_{\mathrm{CoRank}} \gets \frac{1}{|\mathcal{P}|}\sum_{(i,j)\in\mathcal{P}} c_{ij}\log(1+\exp(-y_{ij}d_{ij}))$
        \State $\theta \gets \theta - \eta\nabla_\theta \mathcal{L}_{\mathrm{CoRank}}$
    \EndFor
\EndFor
\Statex \textit{Stage 3. Teacher-free inference}
\For{$t=L,\dots,T'$}
    \State $s_t \gets g_\theta(\mathbf{W}_t^{\mathrm{test}})$
\EndFor
\State \Return $\{s_t\}_{t=L}^{T'}$
\end{algorithmic}
\end{algorithm}

\clearpage
\section{Visualization}
\label{app:score_visualization}

\begin{figure}[h]
\centering
\includegraphics[width=\linewidth]{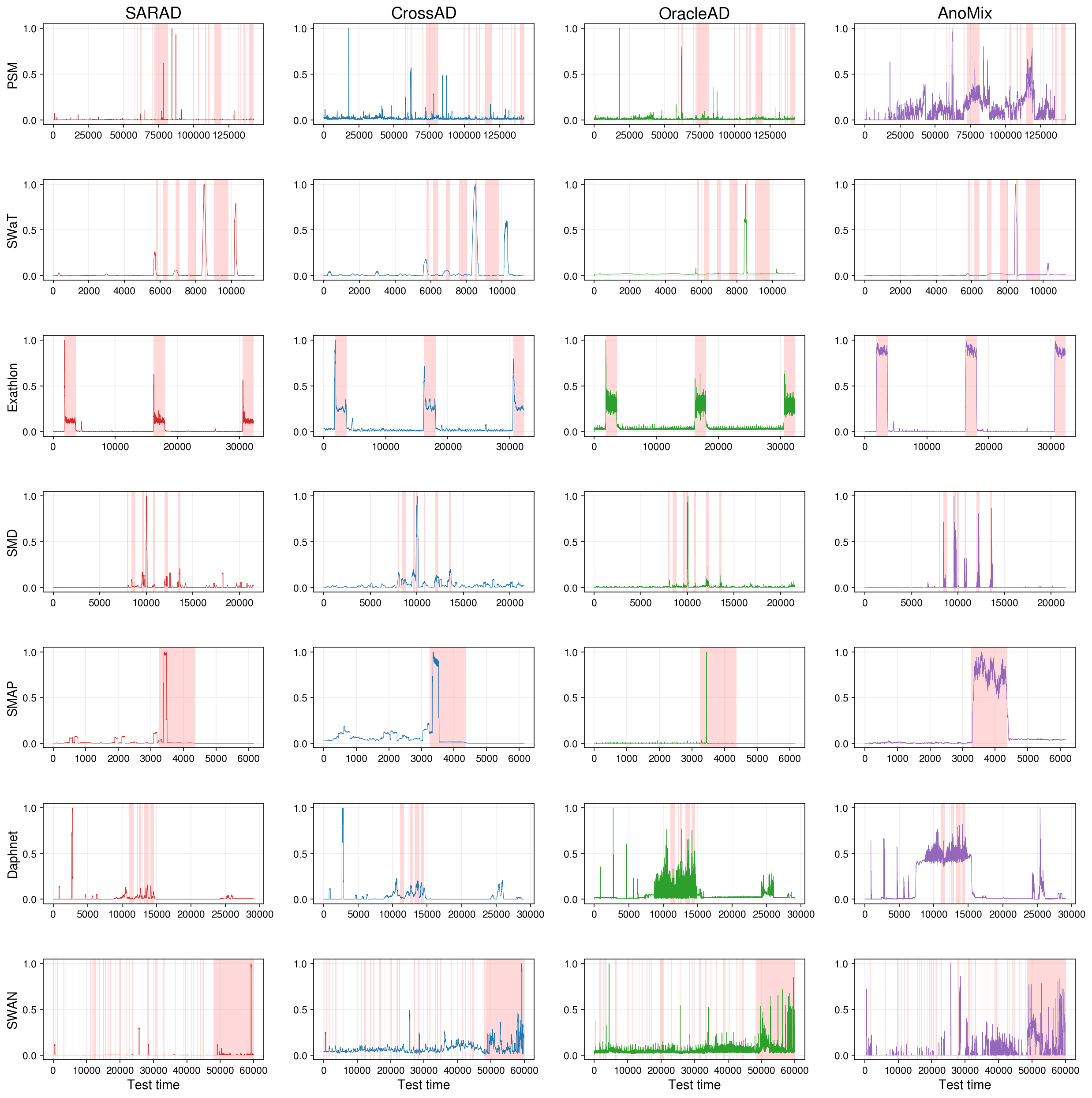}
\caption{\textbf{Test-time anomaly scores.} Rows correspond to the seven benchmark datasets and columns to SARAD, CrossAD, OracleAD, and AnoMix. Scores are min-max normalized to $[0, 1]$. Red shaded bands indicate labeled anomaly segments.}
\label{fig:score_visualization}
\end{figure}

Figure~\ref{fig:score_visualization} visualizes the test-time anomaly scores produced by the three Sel-3 teachers (SARAD, CrossAD, OracleAD) and AnoMix on the seven benchmark datasets.
Each series is min-max normalized to $[0, 1]$ for direct visual comparison, and anomaly segments are shaded in red.
Across these datasets, AnoMix can be compared directly with the teacher score sequences and often assigns high scores to labeled anomaly segments without simply copying any single teacher.

\clearpage
\section{Full Results}
\label{app:full_metrics}

\paragraph{PR Metrics.}
Table~\ref{tab:pr_metrics} reports VUS-PR, AUC-PR, and R-AUC-PR (5-seed mean, \%).

\begin{table*}[h]
\centering
\caption{PR metrics on the seven benchmark datasets (5-seed mean, \%). \textbf{Boldface} indicates the best result in each column, and \underline{underlined} values indicate the second-best result.}
\label{tab:pr_metrics}
\resizebox{\textwidth}{!}{
\begin{tabular}{l|ccc|ccc|ccc|ccc|ccc|ccc|ccc}
\toprule
\multirow{2}{*}{\textbf{Method}} & \multicolumn{3}{c|}{\textbf{PSM}} & \multicolumn{3}{c|}{\textbf{SWaT}} & \multicolumn{3}{c|}{\textbf{Exathlon}} & \multicolumn{3}{c|}{\textbf{SMD}} & \multicolumn{3}{c|}{\textbf{SMAP}} & \multicolumn{3}{c|}{\textbf{Daphnet}} & \multicolumn{3}{c}{\textbf{SWAN}} \\
\cmidrule(lr){2-22}
& \textbf{VUS} & \textbf{AUC} & \textbf{R-AUC} & \textbf{VUS} & \textbf{AUC} & \textbf{R-AUC} & \textbf{VUS} & \textbf{AUC} & \textbf{R-AUC} & \textbf{VUS} & \textbf{AUC} & \textbf{R-AUC} & \textbf{VUS} & \textbf{AUC} & \textbf{R-AUC} & \textbf{VUS} & \textbf{AUC} & \textbf{R-AUC} & \textbf{VUS} & \textbf{AUC} & \textbf{R-AUC} \\
\midrule
Omni & 22.52 & 21.43 & 24.07 & 27.32 & 24.03 & 32.12 & \underline{99.75} & \underline{98.88} & \underline{99.96} & 69.98 & 71.68 & 75.18 & 27.68 & 26.97 & 28.08 & 18.38 & 16.97 & 19.25 & 91.16 & \underline{54.34} & 91.51 \\
USAD\textsuperscript{'20} & 22.73 & 22.52 & 24.35 & 33.39 & 29.60 & 38.56 & 99.63 & 98.47 & 99.93 & 71.97 & 72.42 & 77.36 & 27.58 & 26.86 & 27.98 & 17.15 & 15.24 & 18.31 & 88.60 & 49.50 & 87.86 \\
A.T & 23.13 & 21.95 & 23.18 & 23.55 & 20.94 & 26.87 & 24.88 & 24.59 & 25.28 & 11.80 & 9.74 & 13.74 & 22.30 & 21.35 & 22.90 & 10.33 & 9.40 & 11.35 & 81.07 & 36.05 & 85.01 \\
TranAD\textsuperscript{'22} & 22.25 & 21.27 & 22.61 & \underline{37.86} & \underline{31.38} & \underline{43.86} & 93.88 & 93.35 & 93.97 & 52.05 & 46.92 & 45.51 & 37.06 & 35.94 & 37.54 & 17.98 & 16.48 & 19.29 & 86.48 & 41.25 & 86.87 \\
TimesNet\textsuperscript{'23} & 24.32 & 20.33 & 26.19 & 24.35 & 20.87 & 28.20 & 85.33 & 81.74 & 88.15 & 53.06 & 44.41 & 56.79 & 23.90 & 22.80 & 24.62 & 33.97 & 31.01 & 36.82 & 88.77 & 43.48 & 90.36 \\
SARAD\textsuperscript{'24} & 26.78 & 22.28 & 31.37 & 29.82 & 24.90 & 34.93 & 98.27 & 95.47 & 99.59 & 62.65 & 56.33 & 70.98 & 37.33 & 36.45 & 37.90 & 38.72 & 35.11 & \underline{41.02} & 90.20 & \underline{52.56} & 89.63 \\
MOMENT\textsuperscript{'24} & 16.59 & 15.68 & 17.76 & 20.21 & 17.98 & 22.60 & 97.17 & 93.89 & 98.91 & 10.19 & 8.20 & 12.16 & 27.85 & 27.43 & 28.53 & 35.10 & 31.58 & 37.15 & 86.25 & 44.56 & 86.65 \\
CATCH\textsuperscript{'25} & 23.42 & 19.76 & 27.25 & 24.61 & 20.83 & 28.47 & 97.79 & 94.52 & 99.52 & 67.33 & 60.17 & 76.64 & 27.45 & 26.74 & 27.84 & \underline{38.86} & \underline{35.82} & 40.76 & 89.66 & 44.65 & \underline{91.75} \\
CrossAD\textsuperscript{'25} & 27.66 & 24.44 & \underline{32.01} & 26.33 & 23.70 & 30.82 & 92.47 & 89.33 & 95.76 & \underline{72.52} & \underline{72.48} & \underline{80.65} & 35.62 & 34.27 & 37.42 & 36.31 & 32.67 & 38.99 & \underline{91.66} & 49.04 & \textbf{92.68} \\
OracleAD\textsuperscript{'25} & \underline{28.68} & \underline{27.92} & 29.10 & 26.44 & 23.54 & 29.10 & 97.18 & 96.95 & 97.32 & 52.69 & 48.61 & 55.03 & \underline{40.18} & \underline{39.61} & \underline{40.70} & 34.60 & 31.77 & 37.47 & 86.30 & 48.78 & 86.88 \\
\midrule
\textbf{AnoMix} & \textbf{44.03} & \textbf{54.90} & \textbf{44.75} & \textbf{42.40} & \textbf{38.19} & \textbf{47.25} & \textbf{99.97} & \textbf{99.95} & \textbf{99.98} & \textbf{89.37} & \textbf{81.79} & \textbf{91.99} & \textbf{97.65} & \textbf{96.85} & \textbf{98.40} & \textbf{52.37} & \textbf{47.98} & \textbf{56.29} & \textbf{91.76} & \textbf{58.74} & 90.59 \\
\bottomrule
\end{tabular}}
\end{table*}

\paragraph{ROC Metrics.}
Table~\ref{tab:roc_metrics} reports VUS-ROC, AUC-ROC, and R-AUC-ROC (5-seed mean, \%).

\begin{table*}[h]
\centering
\caption{ROC metrics on the seven benchmark datasets (5-seed mean, \%). \textbf{Boldface} indicates the best result in each column, and \underline{underlined} values indicate the second-best result.}
\label{tab:roc_metrics}
\resizebox{\textwidth}{!}{
\begin{tabular}{l|ccc|ccc|ccc|ccc|ccc|ccc|ccc}
\toprule
\multirow{2}{*}{\textbf{Method}} & \multicolumn{3}{c|}{\textbf{PSM}} & \multicolumn{3}{c|}{\textbf{SWaT}} & \multicolumn{3}{c|}{\textbf{Exathlon}} & \multicolumn{3}{c|}{\textbf{SMD}} & \multicolumn{3}{c|}{\textbf{SMAP}} & \multicolumn{3}{c|}{\textbf{Daphnet}} & \multicolumn{3}{c}{\textbf{SWAN}} \\
\cmidrule(lr){2-22}
& \textbf{VUS} & \textbf{AUC} & \textbf{R-AUC} & \textbf{VUS} & \textbf{AUC} & \textbf{R-AUC} & \textbf{VUS} & \textbf{AUC} & \textbf{R-AUC} & \textbf{VUS} & \textbf{AUC} & \textbf{R-AUC} & \textbf{VUS} & \textbf{AUC} & \textbf{R-AUC} & \textbf{VUS} & \textbf{AUC} & \textbf{R-AUC} & \textbf{VUS} & \textbf{AUC} & \textbf{R-AUC} \\
\midrule
Omni & 51.59 & 59.49 & 52.74 & 59.42 & 59.91 & 63.64 & \underline{99.98} & \underline{99.91} & \textbf{100.00} & 95.40 & 95.17 & 96.34 & 49.93 & 48.90 & 50.83 & 78.60 & 79.20 & 79.01 & 92.06 & \underline{63.19} & 86.54 \\
USAD\textsuperscript{'20} & 52.15 & 58.18 & 52.52 & 68.92 & 70.88 & 72.92 & 99.97 & 99.88 & \underline{99.99} & 95.98 & 95.40 & 96.88 & 49.80 & 48.77 & 50.70 & 76.95 & 76.13 & 78.18 & 89.80 & 56.15 & 81.06 \\
A.T & 53.76 & 54.64 & 44.00 & 56.63 & 54.36 & 59.22 & 37.70 & 37.13 & 38.35 & 44.03 & 42.04 & 44.83 & 40.56 & 39.73 & 41.49 & 46.28 & 44.92 & 47.59 & 86.33 & 50.95 & 78.38 \\
TranAD\textsuperscript{'22} & 55.90 & 55.14 & 48.22 & \underline{77.06} & \underline{72.97} & \textbf{81.54} & 99.50 & 99.47 & 99.50 & 86.51 & 82.75 & 81.84 & 71.96 & 70.85 & 72.81 & 76.71 & 73.04 & 79.00 & 92.02 & 52.60 & 81.50 \\
TimesNet\textsuperscript{'23} & 57.91 & 53.22 & 57.87 & 59.29 & 53.56 & 64.78 & 98.61 & 98.24 & 98.88 & 93.96 & 90.60 & 92.45 & 59.22 & 58.06 & 60.13 & 90.25 & 88.98 & 91.30 & \textbf{93.87} & 56.52 & 86.33 \\
SARAD\textsuperscript{'24} & 58.10 & 57.44 & 62.37 & 68.29 & 64.29 & 72.41 & 99.86 & 99.67 & 99.96 & 95.40 & 93.58 & 96.83 & 75.27 & 74.43 & 75.92 & 89.96 & 88.91 & 90.93 & 92.11 & 58.13 & 84.49 \\
MOMENT\textsuperscript{'24} & 42.39 & 44.07 & 40.29 & 54.95 & 50.25 & 59.33 & 99.66 & 99.24 & 99.83 & 57.27 & 49.95 & 61.94 & 54.72 & 53.35 & 56.10 & \underline{91.71} & \underline{91.03} & \underline{92.28} & 92.07 & 55.78 & 81.22 \\
CATCH\textsuperscript{'25} & 53.99 & 51.92 & 57.53 & 58.82 & 53.47 & 64.30 & 99.83 & 99.59 & 99.96 & 96.45 & 94.40 & \underline{97.89} & 49.69 & 48.66 & 50.58 & 91.37 & 90.89 & 91.87 & 92.16 & 53.23 & \underline{88.17} \\
CrossAD\textsuperscript{'25} & 59.02 & 57.87 & 63.33 & 60.79 & 56.96 & 66.17 & 99.41 & 99.18 & 99.65 & \underline{96.49} & \textbf{97.18} & 97.38 & 52.95 & 51.77 & 54.24 & 89.77 & 89.74 & 90.76 & 92.91 & 59.42 & \textbf{88.37} \\
OracleAD\textsuperscript{'25} & \underline{66.33} & \underline{65.02} & \underline{64.19} & 59.87 & 58.39 & 58.64 & 99.66 & 99.57 & 99.69 & 85.71 & 80.39 & 88.50 & \underline{79.88} & \underline{78.80} & \underline{80.69} & 88.46 & 85.71 & 90.14 & 92.44 & 60.36 & 81.47 \\
\midrule
\textbf{AnoMix} & \textbf{70.24} & \textbf{77.01} & \textbf{69.98} & \textbf{78.56} & \textbf{77.80} & \underline{80.62} & \textbf{100.00} & \textbf{99.99} & \textbf{100.00} & \textbf{98.87} & \underline{97.00} & \textbf{98.81} & \textbf{98.38} & \textbf{97.63} & \textbf{99.02} & \textbf{94.13} & \textbf{92.94} & \textbf{95.07} & \underline{93.75} & \textbf{69.06} & 85.33 \\
\bottomrule
\end{tabular}}
\end{table*}

\paragraph{F1-Based Metrics.}
Table~\ref{tab:f1_metrics} reports point-wise F1, Event-F1~\citep{Evaluation-2021}, and Aff-F~\citep{Local_Evaluation_TSAD-2022} (5-seed mean, \%).

\begin{table*}[h]
\centering
\caption{F1-based metrics on the seven benchmark datasets (5-seed mean, \%). \textbf{Boldface} indicates the best result in each column, and \underline{underlined} values indicate the second-best result.}
\label{tab:f1_metrics}
\resizebox{\textwidth}{!}{
\begin{tabular}{l|ccc|ccc|ccc|ccc|ccc|ccc|ccc}
\toprule
\multirow{2}{*}{\textbf{Method}} & \multicolumn{3}{c|}{\textbf{PSM}} & \multicolumn{3}{c|}{\textbf{SWaT}} & \multicolumn{3}{c|}{\textbf{Exathlon}} & \multicolumn{3}{c|}{\textbf{SMD}} & \multicolumn{3}{c|}{\textbf{SMAP}} & \multicolumn{3}{c|}{\textbf{Daphnet}} & \multicolumn{3}{c}{\textbf{SWAN}} \\
\cmidrule(lr){2-22}
& \textbf{F1} & \textbf{Event} & \textbf{Aff} & \textbf{F1} & \textbf{Event} & \textbf{Aff} & \textbf{F1} & \textbf{Event} & \textbf{Aff} & \textbf{F1} & \textbf{Event} & \textbf{Aff} & \textbf{F1} & \textbf{Event} & \textbf{Aff} & \textbf{F1} & \textbf{Event} & \textbf{Aff} & \textbf{F1} & \textbf{Event} & \textbf{Aff} \\
\midrule
Omni & 32.33 & 33.23 & 71.17 & 36.08 & 42.27 & 71.63 & \underline{98.90} & \textbf{100.00} & \underline{99.92} & 65.77 & 73.18 & 83.23 & 40.81 & \textbf{100.00} & 94.05 & 31.66 & 33.39 & 86.13 & 49.72 & 49.17 & 71.05 \\
USAD\textsuperscript{'20} & 31.42 & 32.11 & 71.52 & 45.84 & \underline{44.86} & 74.07 & 98.70 & \underline{99.85} & 99.91 & 64.51 & 66.15 & 83.88 & 40.81 & \textbf{100.00} & 94.05 & 30.87 & 31.11 & 85.85 & 49.17 & 49.17 & 71.05 \\
A.T & 30.13 & 28.58 & 66.79 & 32.80 & 42.83 & 70.23 & 29.57 & 78.63 & 87.28 & 15.07 & 35.43 & 64.33 & 30.92 & \underline{57.50} & 77.16 & 16.68 & 46.50 & 79.35 & 49.17 & 32.23 & 53.48 \\
TranAD\textsuperscript{'22} & 29.30 & 29.20 & 69.43 & \underline{48.65} & 33.39 & 71.64 & 97.62 & \textbf{100.00} & 99.11 & 52.64 & \underline{76.91} & \textbf{91.88} & \underline{63.20} & \textbf{100.00} & 93.14 & 29.36 & 61.49 & 88.35 & 49.17 & 49.17 & 71.05 \\
TimesNet\textsuperscript{'23} & 29.20 & 29.20 & 69.43 & 30.39 & 30.26 & 71.63 & 90.08 & 94.00 & 97.45 & 51.94 & 67.34 & 86.40 & 40.20 & \textbf{100.00} & 92.58 & 40.48 & 58.10 & 82.44 & 49.17 & 49.17 & 71.05 \\
SARAD\textsuperscript{'24} & 31.10 & 29.20 & 69.43 & 36.93 & 35.53 & 71.64 & 97.77 & 98.92 & 99.79 & 55.60 & 65.40 & 89.31 & 52.73 & \textbf{100.00} & 94.05 & 47.63 & 58.84 & \underline{90.81} & 49.17 & 49.17 & 71.05 \\
MOMENT\textsuperscript{'24} & 29.20 & 29.20 & 69.43 & 30.05 & \textbf{49.88} & \underline{76.89} & 94.78 & 98.83 & 99.72 & 15.09 & 22.12 & 71.66 & 30.90 & \textbf{100.00} & 93.75 & 46.99 & 53.02 & 89.70 & 49.17 & 49.17 & 71.05 \\
CATCH\textsuperscript{'25} & 29.36 & 39.75 & \underline{78.87} & 31.60 & 40.69 & \textbf{78.71} & 98.26 & 98.95 & 99.81 & 57.37 & 70.97 & \underline{91.43} & 40.80 & \textbf{100.00} & 94.05 & 46.15 & 54.07 & 89.46 & 49.17 & 49.17 & 71.05 \\
CrossAD\textsuperscript{'25} & 29.83 & 43.74 & \textbf{81.07} & 33.26 & 37.96 & 75.96 & 97.71 & 97.76 & 99.60 & \underline{68.21} & 67.25 & 88.26 & 40.82 & \textbf{100.00} & \underline{94.75} & \underline{49.03} & 52.18 & 84.40 & 49.18 & 49.17 & 71.05 \\
OracleAD\textsuperscript{'25} & \underline{35.71} & \underline{45.93} & 77.66 & 35.37 & 37.15 & 74.98 & 95.34 & \textbf{100.00} & 99.59 & 48.00 & \textbf{88.26} & 89.88 & 56.92 & \textbf{100.00} & 93.42 & 41.13 & \underline{68.14} & 90.11 & \underline{50.05} & \underline{49.18} & \underline{71.05} \\
\midrule
\textbf{AnoMix} & \textbf{56.69} & \textbf{50.93} & 70.97 & \textbf{53.43} & 42.89 & 74.58 & \textbf{99.36} & \textbf{100.00} & \textbf{100.00} & \textbf{82.36} & 72.85 & 85.84 & \textbf{96.82} & \textbf{100.00} & \textbf{99.97} & \textbf{50.07} & \textbf{78.21} & \textbf{94.28} & \textbf{53.57} & \textbf{50.13} & \textbf{71.12} \\
\bottomrule
\end{tabular}}
\end{table*}

\clearpage
\section{Datasets}
\label{app:datasets}

Table~\ref{tab:datasets} summarizes the seven multivariate time series anomaly detection datasets used in the paper.
We selected these datasets to evaluate AnoMix across diverse benchmark conditions, including different training-set sizes, anomaly ratios, event counts, and numbers of variates.
We treat subsets from the same benchmark source as separate benchmark instances rather than different splits of a single dataset, since they differ in dimensionality and data distribution, following prior works~\citep{OmniAnomaly-2019,D3R-2023,CAROTS-2025}.
For datasets with multiple available subsets, the subset was fixed before evaluating AnoMix by choosing among subsets with sufficient anomaly ratios for evaluation.
The first six are sourced from TSB-AD~\citep{Elephant_in_the_Room-2024}, which provides standardized preprocessing, train/test splits, and anomaly labels, and SWAN is sourced from TAB~\citep{TAB-2025}.

\begin{table}[h]
\centering
\caption{Statistics of the seven benchmark datasets.}
\label{tab:datasets}
\footnotesize
\setlength{\tabcolsep}{3.2pt}
\resizebox{\textwidth}{!}{
\begin{tabular}{c|clcccccc}
\toprule
Dataset & Subset & Domain & $D$ & Train & Validation & Test & Anomaly (\%) & Events \\
\midrule
PSM      & --   & Server monitoring             & 25 & 67{,}500 & 7{,}500 & 142{,}624 & 11.2 & 72 \\
SWaT     & --   & Cyber-physical control        & 66 &  3{,}374 &   375 &  11{,}247 & 13.2 & 6 \\
Exathlon & id22 & Distributed stream processing & 31 &  9{,}689 & 1{,}077 &  32{,}300 & 12.6 & 3 \\
SMD      & id6  & Internet-company server       & 38 &  6{,}462 &   718 &  21{,}542 &  6.1 & 10 \\
SMAP     & id1  & Satellite telemetry           & 25 &  1{,}846 &   206 &   6{,}157 & 13.7 & 1 \\
Daphnet  & --   & Wearable accelerometer        &  9 &  8{,}723 &   970 &  29{,}081 &  5.9 & 6 \\
SWAN     & --   & Solar-flare monitoring        & 38 & 54{,}000 & 6{,}000 &  60{,}000 & 32.6 & 6{,}148 \\
\bottomrule
\end{tabular}}
\end{table}

\begin{itemize}
    \item \textbf{PSM}~\citep{PSM-2021}: Server-monitoring metrics collected from eBay's production infrastructure. The 25 channels cover CPU, memory, disk, and network usage, with anomalies corresponding to real software incidents.

    \item \textbf{SWaT}~\citep{SWaT-2016}: Sensor and actuator readings from the Secure Water Treatment testbed. The 66 channels cover flow, tank-level, valve, and chemical-dosing signals under cyber-physical attacks. We use the processed TSB-AD version, while the original SWaT data are distributed by iTrust and require an access request through the official dataset website before download: \url{https://itrust.sutd.edu.sg/itrust-labs_datasets/dataset_info/}.

    \item \textbf{Exathlon}~\citep{Exathlon-2021}: Application-level metrics from distributed stream-processing jobs running on Apache Spark. The 31 channels include job latency, throughput, CPU utilization, and garbage-collection statistics, and anomalies are caused by resource contention, configuration errors, and infrastructure failures.

    \item \textbf{SMD}~\citep{OmniAnomaly-2019}: Server Machine Dataset consisting of 38 monitoring metrics (CPU, memory, disk, network) collected from a large Internet company.

    \item \textbf{SMAP}~\citep{SMAP_MSL-2018}: Telemetry data from NASA's Soil Moisture Active Passive satellite. The 25 channels include sensor readings and actuator commands, with anomalies corresponding to sensor malfunctions and actuator failures.

    \item \textbf{Daphnet}~\citep{Daphnet-2010}: Wearable accelerometer data from Parkinson's disease patients. Three tri-axial accelerometers produce 9 channels, and anomalies correspond to freezing-of-gait episodes.

    \item \textbf{SWAN}~\citep{SWAN-2020}: SHARP magnetic-field metrics from NASA SDO/HMI used for solar flare monitoring. The 38 channels describe magnetic-field properties, region geometry, and X-ray flux history, with anomalies corresponding to active regions associated with elevated flare activity.
\end{itemize}

\clearpage
\section{Metrics}
\label{app:metrics}

We report threshold-independent PR and ROC metrics in three forms.
The threshold-independent metrics capture complementary aspects of TSAD performance, including pointwise discrimination, tolerance to small temporal offsets, and ranking quality across the full score range.

They summarize performance across all possible score thresholds and therefore evaluate the quality of the anomaly scores themselves rather than one specific threshold choice.
\begin{itemize}
    \item \textbf{AUC-PR, AUC-ROC}~\citep{PR_ROC-2006,ROC-2006}: Standard pointwise areas under the Precision-Recall and ROC curves. AUC-PR is especially informative under severe class imbalance, while AUC-ROC summarizes global ranking quality over normal and anomalous points.
    \item \textbf{R-AUC-PR, R-AUC-ROC}~\citep{PrecRec_for_TS-2018,VUS-2022}: Range-based variants that relax exact timestamp matching and instead credit predictions that fall within a tolerance window around each anomaly segment. They are therefore more appropriate when slight temporal delays or advances should not be penalized as harshly as pointwise errors.
    \item \textbf{VUS-PR, VUS-ROC}~\citep{VUS-2022}: Volume Under the Surface extends range-based AUC by integrating not only over score thresholds but also over tolerance-window sizes. VUS therefore provides a broader threshold-free view of time-series detection quality than either standard or fixed-range AUC.
\end{itemize}

\clearpage
\section{Baselines and Experimental Protocol}
\label{app:baselines}

\paragraph{Deep TSAD Baselines.}
\begin{itemize}
    \item \textbf{OmniAnomaly} (KDD 2019)~\citep{OmniAnomaly-2019}: a stochastic reconstruction model that combines recurrent dynamics with a variational latent-variable formulation for sequence anomaly detection.
    \item \textbf{USAD} (KDD 2020)~\citep{USAD-2020}: an unsupervised adversarial autoencoder that detects anomalies through reconstruction inconsistency under dual decoder objectives.
    \item \textbf{Anomaly Transformer} (ICLR 2022)~\citep{AnomalyTransformer-2021}: a Transformer-based detector that uses association discrepancy to identify unusual attention structure.
    \item \textbf{TranAD} (VLDB 2022)~\citep{TranAD-2022}: a forecasting-oriented Transformer that predicts future behavior and flags deviations from those predictions.
    \item \textbf{TimesNet} (ICLR 2023)~\citep{TimesNet-2022}: a general time-series backbone that models multi-period temporal variation and is used here as a forecasting-style anomaly detector.
    \item \textbf{MOMENT} (ICML 2024)~\citep{MOMENT-2024}: a pretrained time-series foundation model used in our target-dataset benchmark as a foundation-model-based anomaly baseline. For Table~\ref{tab:main}, we follow the TSB-AD wrapper convention~\citep{Elephant_in_the_Room-2024}: the pretrained MOMENT backbone is kept frozen and a lightweight reconstruction head is fine-tuned for $2$ epochs on each target training split.
    \item \textbf{SARAD} (NeurIPS 2024)~\citep{SARAD-2024}: a spatial association-aware TSAD method that detects anomalies through reconstruction in association space.
    \item \textbf{CATCH} (ICLR 2025)~\citep{CATCH-2025}: a frequency-patching TSAD framework that reconstructs spectral patterns while modeling channel interactions.
    \item \textbf{CrossAD} (NeurIPS 2025)~\citep{CrossAD-2025}: a multi-scale detector that models cross-scale associations and cross-window temporal context.
    \item \textbf{OracleAD} (NeurIPS 2025)~\citep{OracleAD-2025}: a deviation-based detector that scores anomalies by measuring departure from learned normality.
\end{itemize}

\paragraph{Score-Ensemble Baselines.}
\begin{itemize}
    \item \textbf{Mean}, \textbf{Max}, and \textbf{Median}: pointwise aggregation of normalized teacher scores using fixed reduction rules.
    \item \textbf{Rank Fusion}: conversion of each teacher score sequence into percentile ranks, followed by averaging of those ranks.
    \item \textbf{RRF} (SIGIR 2009)~\citep{RRF-2009}: aggregation of teacher rankings through reciprocal-rank weighting.
\end{itemize}

\paragraph{Zero-Shot Transfer Baselines.}
\begin{itemize}
    \item \textbf{MOMENT} (ICML 2024)~\citep{MOMENT-2024}: evaluated in the zero-shot transfer experiment without fitting any component on the held-out target dataset.
    \item \textbf{MOIRAI} (ICML 2024)~\citep{MOIRAI-2024}: a universal pretrained forecasting model designed for broad transfer across domains and sampling regimes.
    \item \textbf{TimesFM} (ICML 2024)~\citep{TimesFM-2024}: a decoder-style foundation model trained primarily for large-scale forecasting transfer.
    \item \textbf{Lag-Llama} (NeurIPS 2023 Workshop)~\citep{LagLlama-2024}: a generative foundation model for probabilistic time-series forecasting.
    \item \textbf{Chronos} (TMLR 2024)~\citep{Chronos-2024}: a model that casts time-series forecasting into a language-modeling-style token prediction problem.
    \item \textbf{DADA} (ICLR 2025)~\citep{DADA-2025}: a TSAD-specific foundation model pretrained for zero-shot anomaly detection with adaptive bottlenecks and dual adversarial decoders.
\end{itemize}

\paragraph{Code Bases.}
\begin{itemize}
    \item \textbf{TSB-AD}: \url{https://github.com/TheDatumOrg/TSB-AD}
    \item \textbf{CATCH}: \url{https://github.com/decisionintelligence/CATCH}
    \item \textbf{SARAD}: \url{https://github.com/daidahao/SARAD}
    \item \textbf{CrossAD}: \url{https://github.com/decisionintelligence/CrossAD}
    \item \textbf{MOIRAI}: \url{https://github.com/SalesforceAIResearch/uni2ts}
    \item \textbf{Lag-Llama}: \url{https://github.com/time-series-foundation-models/lag-llama}
    \item \textbf{DADA}: \url{https://github.com/iambowen/DADA}
\end{itemize}

\paragraph{Shared Data and Evaluation Controls.}
We reran all ten target-dataset baselines and teachers on the same dataset files and fixed train/test splits; no published benchmark values were reused.
Five-run comparisons use seeds $\{42,123,456,789,1024\}$, and training uses no anomaly labels.
All scores use the common TSB-AD metric implementation with evaluation window $100$ for range-based and VUS metrics, independently of the model's input-window length.

\paragraph{Model-Specific Settings and Optimization.}
The TSB-AD implementations and local adapters use recommended baseline configurations without per-dataset hyperparameter searches.
Table~\ref{tab:baseline_protocol} summarizes the target-dataset benchmark configuration; windowing, normalization, epoch limits, and checkpoint restoration remain implementation-specific.
Except for MOMENT's fixed adaptation schedule, trainable baselines and teachers use the last $10\%$ of the training split for validation and early stopping with patience $3$.
OmniAnomaly, TimesNet, SARAD, CATCH, and OracleAD share a training loop with at most $10$ epochs, patience $3$, and restoration of the lowest-validation-loss checkpoint.
USAD, Anomaly Transformer, TranAD, and CrossAD use their wrappers' stopping and checkpoint rules.
For Table~\ref{tab:main}, MOMENT fine-tunes its reconstruction head for a fixed $2$ epochs and uses the final checkpoint, without early stopping; its pretrained backbone remains frozen.
In Table~\ref{tab:zero_shot}, MOMENT instead uses the pretrained checkpoint without any target-dataset fine-tuning.
AnoMix uses the final checkpoint after $10$ epochs; Appendix~\ref{app:implementation_details} gives its per-dataset settings and training-loss-based learning-rate choice.

\begin{table}[h]
\centering
\caption{Baseline configurations for Table~\ref{tab:main}.}
\label{tab:baseline_protocol}
\small
\setlength{\tabcolsep}{5pt}
\begin{tabular}{l|ccp{0.43\textwidth}}
\toprule
Model & Window & Learning rate & Stopping rule \\
\midrule
OmniAnomaly & 100 & $2\!\times\!10^{-3}$ & Early stopping \\
USAD & 100 & $10^{-3}$ & Early stopping \\
Anomaly Transformer & 50 & $10^{-3}$ & Early stopping \\
TranAD & 10 & $10^{-3}$ & Early stopping \\
TimesNet & 96 & $10^{-4}$ & Early stopping \\
SARAD & 100 & $7\!\times\!10^{-4}$ & Early stopping \\
MOMENT & 64 & $10^{-4}$ & Fixed 2 epochs (head fine-tuning) \\
CATCH & 100 & $10^{-4}$ & Early stopping \\
CrossAD & 192 & $10^{-4}$ & Early stopping \\
OracleAD & 10 & $10^{-3}$ & Early stopping \\
\bottomrule
\end{tabular}
\end{table}

\paragraph{Teacher Checkpoints and Student Seeds.}
For each dataset and teacher, a seed from the common seed set is fixed in advance, without performance-based seed selection.
Its checkpoint supplies cached training scores, min--max normalized using training-score statistics and reused across five student seeds.
Student variability in Appendix~\ref{app:significance} is conditional on this fixed supervision and excludes teacher-retraining variability.
Each standalone teacher row in Table~\ref{tab:main} instead averages five teacher runs.

\paragraph{Fixed-Checkpoint Transfer Baselines.}
The six foundation baselines in Table~\ref{tab:zero_shot} use fixed released checkpoints and model-specific preprocessing without target-dataset weight updates.
Only AnoMix-ZS is trained on the source datasets with five seeds; Appendix~\ref{app:significance} reports its standard deviations.

\clearpage
\section{Teacher Sets and Robustness}
\label{app:teacher_selection}

\subsection{Teacher-Set Selection}
\label{app:teacher_set_selection}

The main teacher-set study considers three types of configuration.
\emph{Sel-3} is the fixed benchmark teacher set used by AnoMix in the main experiments and consists of SARAD, CrossAD, and OracleAD.
\emph{LOO Best-3} denotes leave-one-out teacher selection, where the target dataset is excluded and teachers are ranked only by their average performance on the remaining six datasets in the seven-task analysis.
\emph{All-10} uses the entire teacher pool without selection.
The purpose of this comparison is not to derive a universal teacher-selection rule, but to clarify what the fixed benchmark choice represents and how sensitive AnoMix is to alternative teacher pools.

The fixed \emph{Sel-3} configuration was chosen before evaluating AnoMix for analysis convenience and is intended as a compact and representative benchmark setting rather than a claimed optimum.
It is formed by taking the three most recent state-of-the-art TSAD detectors in the teacher pool and therefore combines SARAD from association-space reconstruction, CrossAD from cross-scale association modeling, and OracleAD from deviation-based scoring.
This choice keeps the training-time teacher pool small while still covering heterogeneous anomaly-scoring behavior.
It also matches the design philosophy of AnoMix, which aims to distill comparative supervision from complementary teachers into a single deployable student.

The leave-one-out analysis is useful because it removes target-dataset information from teacher selection.
The resulting per-target ranking is shown in Table~\ref{tab:loo_ranking}.
Under this protocol, SARAD and CrossAD occupy the top two positions on six of the seven target datasets, indicating that they are the most reliable cross-dataset teachers in the candidate pool.
The third position is typically taken by OracleAD or CATCH.
This pattern helps explain why the fixed benchmark three-teacher set remains competitive even when compared with teacher sets obtained from target-excluded ranking.

\begin{table}[h]
\centering
\caption{LOO teacher ranking on the seven benchmark datasets. For each target dataset, teachers are ranked by average VUS-PR (\%) on the other six datasets. Omni denotes OmniAnomaly, and A.T denotes Anomaly Transformer. \textbf{Boldface} indicates the top three teachers for each target dataset.}
\label{tab:loo_ranking}
\small
\setlength{\tabcolsep}{3pt}
\begin{tabular}{c|ccccccc}
\toprule
Rank & PSM & SWaT & Exathlon & SMD & SMAP & Daphnet & SWAN \\
\midrule
1 & \textbf{SARAD} & \textbf{CrossAD} & \textbf{CrossAD} & \textbf{SARAD} & \textbf{CrossAD} & \textbf{CrossAD} & \textbf{SARAD} \\
2 & \textbf{CrossAD} & \textbf{SARAD} & \textbf{SARAD} & \textbf{OracleAD} & \textbf{SARAD} & \textbf{SARAD} & \textbf{CrossAD} \\
3 & \textbf{CATCH} & \textbf{CATCH} & \textbf{CATCH} & \textbf{CrossAD} & \textbf{CATCH} & \textbf{USAD} & \textbf{OracleAD} \\
4 & USAD & OracleAD & OracleAD & CATCH & USAD & Omni & CATCH \\
5 & OracleAD & Omni & USAD & TranAD & Omni & OracleAD & USAD \\
6 & Omni & USAD & Omni & USAD & OracleAD & CATCH & Omni \\
7 & TranAD & TranAD & TranAD & Omni & TranAD & TranAD & TranAD \\
8 & TimesNet & TimesNet & TimesNet & MOMENT & TimesNet & TimesNet & TimesNet \\
9 & MOMENT & MOMENT & MOMENT & TimesNet & MOMENT & MOMENT & MOMENT \\
10 & A.T & A.T & A.T & A.T & A.T & A.T & A.T \\
\bottomrule
\end{tabular}
\end{table}

\subsection{Teacher-Pool Robustness}
\label{app:teacher_pool_robustness}

We stress-test teacher quality and score-sequence homogeneity on all seven datasets using five student seeds, with the same TCN, CoRank objective, and training protocol as the main experiment.
HML-3 combines SARAD, OmniAnomaly, and Anomaly Transformer, spanning high, middle, and low performance tiers.
Mid-3 contains USAD, OracleAD, and OmniAnomaly, while Bottom-3 contains TimesNet, MOMENT, and Anomaly Transformer.
These pools remain fixed across datasets.
Corr-3 selects the three-teacher subset with the highest mean pairwise Spearman correlation on each dataset's unlabeled training-score sequences, using neither anomaly labels nor target test scores.
For each pool, we compare AnoMix with the best constituent teacher and the normalized mean-score ensemble built from the same teacher checkpoints.

Table~\ref{tab:teacher_pool_stress} reports macro averages over the seven datasets.
The standard deviation is computed across the five seed-level macro averages.
Mean Spearman measures pairwise teacher correlation on training scores, and pair support is the fraction of candidate training pairs satisfying $\max_m|\Delta_{ij}^{(m)}|>\delta$.
Gains are averaged against each dataset's best constituent teacher and mean ensemble.

\begin{table}[h]
\centering
\caption{Teacher-pool stress test. VUS-PR is reported as mean $\pm$ standard deviation (\%), and gains are in \%pt.}
\label{tab:teacher_pool_stress}
\small
\setlength{\tabcolsep}{4pt}
\begin{tabular}{l|ccccc}
\toprule
Teacher set & Mean Spearman & Pair support & AnoMix VUS-PR & vs.\ best teacher & vs.\ mean ensemble \\
\midrule
Sel-3 & 0.463 & 36.8\% & $\mathbf{73.93\pm0.51}$ & $+15.49$ & $+16.95$ \\
HML-3 & 0.248 & 27.1\% & $66.17\pm1.27$ & $+9.65$ & $+14.11$ \\
Mid-3 & 0.607 & 33.9\% & $68.94\pm0.57$ & $+10.09$ & $+15.15$ \\
Bottom-3 & 0.024 & 3.9\% & $48.74\pm3.40$ & $-2.17$ & $-1.14$ \\
Corr-3 & 0.844 & 27.3\% & $69.22\pm0.70$ & $+14.67$ & $+15.54$ \\
\bottomrule
\end{tabular}
\end{table}

AnoMix exceeds both comparators with mixed-quality, medium-quality, and highly correlated pools.
HML-3 improves on its best teacher by $9.65$\%pt despite combining detectors from different quality tiers.
Bottom-3, however, provides only $3.9\%$ pair support and falls below both comparators, exposing a failure case with weak teachers and sparse pair coverage.
Sel-3 also provides broader pair support than Corr-3 and achieves higher student performance.
This is consistent with complementary teachers supplying informative gaps on different pairs, although these comparisons do not isolate the effects of teacher quality and coverage.
High rank correlation does not imply identical score gaps, so Corr-3 can still supply distinct pairwise signals.

Together with the target-excluded LOO analysis and All-10 control, these results broaden the evidence beyond the fixed Sel-3 pool.
The tested pools are not an average over random subsets, and their success does not establish robustness to shared systematic teacher errors.
Agreement based only on teacher outputs cannot distinguish unanimously correct preferences from unanimously incorrect ones, as discussed in Appendix~\ref{app:limitations}.

\clearpage
\section{Implementation Details}
\label{app:hparams}
\label{app:implementation_details}

Every reported run uses the single-head TCN student described in Appendix~\ref{app:student_architectures}.
The student has hidden dimension $H\!=\!64$, six causal blocks with dilations $\{1,2,4,8,16,32\}$, kernel size $k\!=\!3$, dropout $0.1$, and under 80k parameters on the seven benchmark settings.
Its receptive field is 127 time steps, and most datasets therefore use a window size of 127 in order to fully use the available temporal context.
Student training uses only the training split and no anomaly labels.
We train for 10 epochs with AdamW, weight decay $10^{-4}$, cosine annealing, and a CoRank score-gap threshold $\delta\!=\!0.1$.
Within each mini-batch, CoRank evaluates all unordered non-identical time-point pairs and excludes reverse duplicates.
Student results use five random seeds under fixed data splits and fixed teacher-score caches, so their standard deviations measure student optimization variability conditional on the teacher supervision.
The zero-shot foundation baselines use fixed released checkpoints rather than five independently trained models.
Table~\ref{tab:hparams} lists the remaining per-dataset choices.

We select per-dataset learning rates based on training-loss stability, choosing the rate at which the CoRank loss decreases gradually rather than collapsing to near zero within a few epochs.
For SMD and Daphnet, the loss converges rapidly at the default $10^{-3}$, so we use a lower learning rate of $10^{-5}$.

\begin{table}[h]
\centering
\caption{Per-dataset student hyperparameters.}
\label{tab:hparams}
\small
\begin{tabular}{l|ccc}
\toprule
Dataset & Learning Rate & Window Size & Batch Size \\
\midrule
PSM          & $10^{-3}$ & 127 & 64 \\
SWaT         & $10^{-3}$ & 127 & 64 \\
Exathlon     & $10^{-3}$ & 127 & 64 \\
SMD          & $10^{-5}$ & 64  & 64 \\
SMAP         & $10^{-3}$ & 127 & 64 \\
Daphnet      & $10^{-5}$ & 127 & 64 \\
SWAN         & $10^{-3}$ & 127 & 64 \\
\bottomrule
\end{tabular}
\end{table}

\begin{figure}[h]
    \centering
    \includegraphics[width=\textwidth]{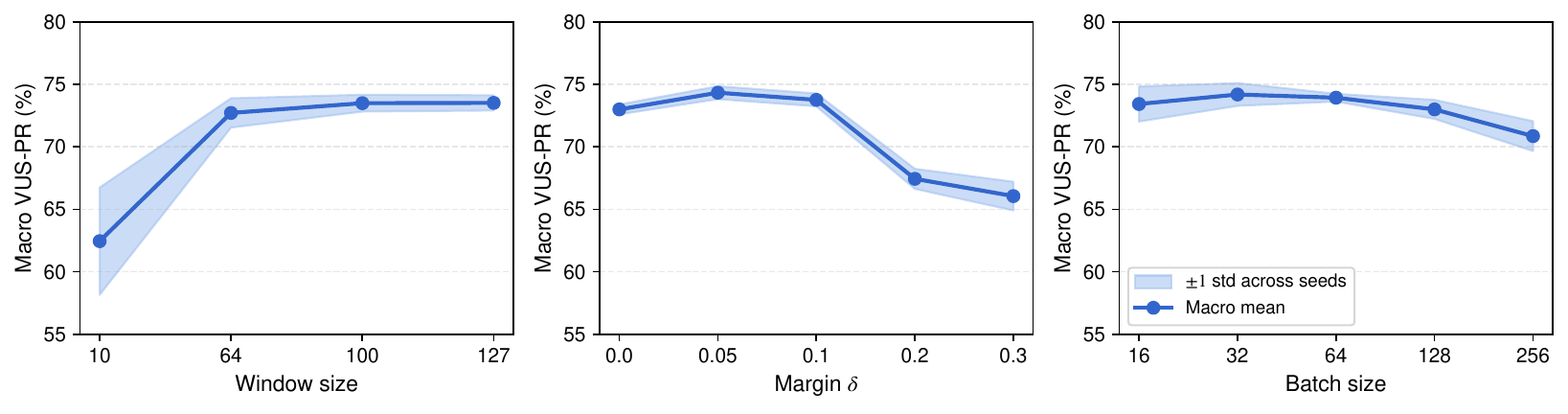}
    \caption{Macro-average hyperparameter sensitivity across the seven datasets. The line denotes the macro mean across datasets, and the shaded band denotes the $\pm 1$ standard deviation across five random seeds.}
    \label{fig:hp_sensitivity}
\end{figure}

Figure~\ref{fig:hp_sensitivity} reports hyperparameter sensitivity with respect to window size $L$, score-gap threshold $\delta$, and batch size.
Window size and batch size are largely flat across the moderate values used in practice, but performance collapses at $L\!=\!10$, where the receptive field is too short to reflect meaningful temporal context.
The score-gap threshold shows the most informative pattern, with small positive values essentially tied at the optimum, while $\delta \ge 0.2$ drops sharply by roughly seven points.
This indicates that the threshold acts as a meaningful filter on small teacher score gaps.
Small positive thresholds retain informative comparisons while suppressing ambiguous pairs, whereas larger thresholds discard too many teacher-derived comparisons and noticeably weaken supervision.

The reported runs use the dataset-specific defaults in Table~\ref{tab:hparams}.
For a new dataset, we recommend choosing within the near-optimal ranges $L \in [64, 127]$, $\delta \in [0.05, 0.1]$, and $\mathrm{bs} \in [32, 64]$, where each setting stays within roughly 1\%pt of the macro optimum across the seven datasets.

\clearpage
\section{Optimization}
\label{app:training_stability}

\begin{figure}[h]
    \centering
    \includegraphics[width=\textwidth]{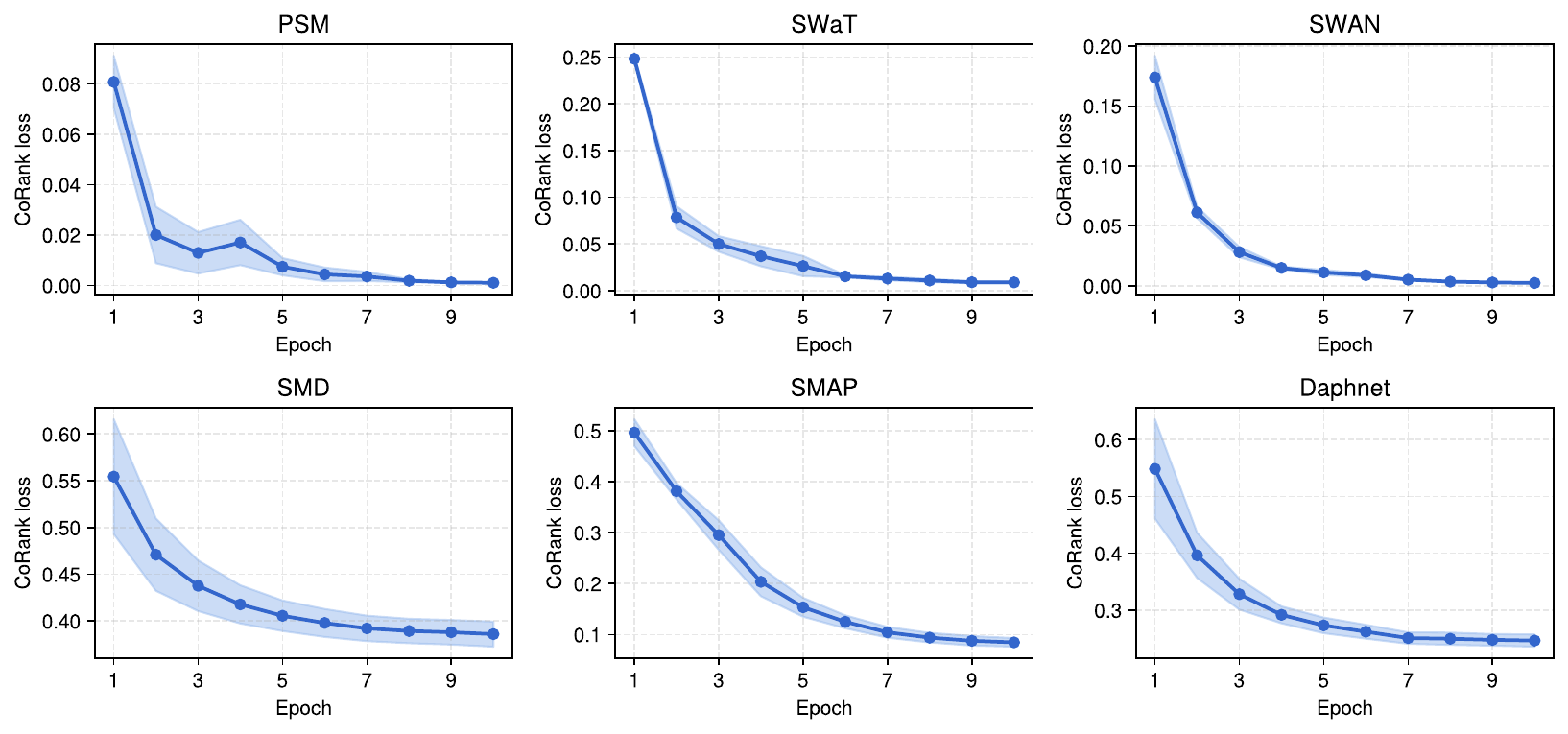}
    \caption{AnoMix CoRank training loss over 10 epochs on each of the six benchmark datasets. The line denotes the five-seed mean, and the shaded region denotes the $\pm 1$ standard deviation across five random seeds.}
    \label{fig:train_loss}
\end{figure}

Figure~\ref{fig:train_loss} shows the per-epoch CoRank training loss for each of the original six benchmark datasets, averaged over five random seeds, with the shaded region denoting the $\pm 1$ standard deviation across five random seeds.
All six curves decrease monotonically under the shared 10-epoch budget, and the across-seed band is narrow on every dataset.
Some datasets show faster early decay than others, but the overall training behavior is stable across these tasks.

\clearpage
\section{Model Complexity and Inference Efficiency}
\label{app:efficiency}

Table~\ref{tab:efficiency} compares model complexity, parameter count, peak GPU memory during single-sample inference, and FLOPs per forward pass on a standardized synthetic input ($L{=}100$, $D{=}200$).
We use a common input size rather than a specific benchmark to isolate architectural cost from per-dataset feature count.
Peak memory is measured with \texttt{torch.cuda.max\_memory\_allocated} after warmup.
We estimate FLOPs with \texttt{fvcore} when tracing is supported and use \texttt{ptflops} estimates for LSTM modules that \texttt{fvcore} cannot trace.
The complexity column drops fixed model hyperparameters such as hidden dimension and reports scaling only in the data-side quantities.
AnoMix requires only $8.65$M FLOPs per window and $9.59$ MB of peak GPU memory at inference, corresponding to $127{,}000\times$ fewer FLOPs than MOMENT and $93\times$ lower peak memory than OracleAD.

These measurements describe deployment-time inference with the student alone.
They exclude the offline costs of training or obtaining the teachers, generating and caching their training scores, and training the student.
AnoMix therefore requires offline teacher computation; the reported inference savings do not imply lower end-to-end training cost than training a single detector.

\begin{table}[h]
\centering
\caption{Model complexity and single-model inference cost on synthetic input ($L{=}100$, $D{=}200$, batch{=}1), excluding offline training and teacher scoring. $L$=window, $D$=features, $P$=patch count.}
\label{tab:efficiency}
\small
\begin{tabular}{c|cccc}
\toprule
Model & Complexity & Params & Memory (MB) & FLOPs \\
\midrule
SARAD & $O(L^2D)$ & 9.66M & 58.04 & 2.03G \\
CrossAD & $O(L^2D)$ & 2.64M & 178.49 & 8.81G \\
OracleAD & $O(LD\!+\!D^2)$ & 211.72M & 893.94 & 21.21G \\
CATCH & $O(P^2D)$ & 13.52M & 86.17 & 1.18G \\
Omni & $O(LD)$ & 314k & 27.55 & 31.64M \\
USAD & $O(LD)$ & 1.00M & 13.82 & 1.60M \\
A.T & $O(L^2D)$ & 5.16M & 43.10 & 545.33M \\
TranAD & $O(L^2D)$ & 3.33M & 37.50 & 237.72M \\
TimesNet & $O(L\!\log\!L\!+\!LD)$ & 80k & 9.90 & 22.90M \\
MOMENT & $O(L^2D)$ & 109.64M & 1035.50 & 1.10T \\
\midrule
\textbf{AnoMix} & $O(LD)$ & \textbf{87k} & \textbf{9.59} & \textbf{8.65M} \\
\bottomrule
\end{tabular}
\end{table}

All reported efficiency measurements and student training runs were performed on a single local workstation.
Table~\ref{tab:system_config} summarizes the system configuration.

\begin{table}[h]
\centering
\caption{System configuration used for the reported experiments.}
\label{tab:system_config}
\small
\begin{tabular}{c|c}
\toprule
Component & Configuration \\
\midrule
CPU & Intel Core Ultra 7 265K, 20 cores \\
GPU & NVIDIA GeForce RTX 5090, 32 GB memory \\
System memory & 96 GB RAM \\
Operating system & Ubuntu 24.04.4 LTS, Linux 6.14 \\
NVIDIA driver & 570.153.02 \\
Software stack & Python 3.13, PyTorch 2.7.0+cu128, CUDA runtime 12.8 \\
\bottomrule
\end{tabular}
\end{table}

\clearpage
\newpage
\section{Additional Ablations}
\label{app:additional_ablations}

\begin{figure}[h]
    \centering
    \includegraphics[width=0.88\textwidth]{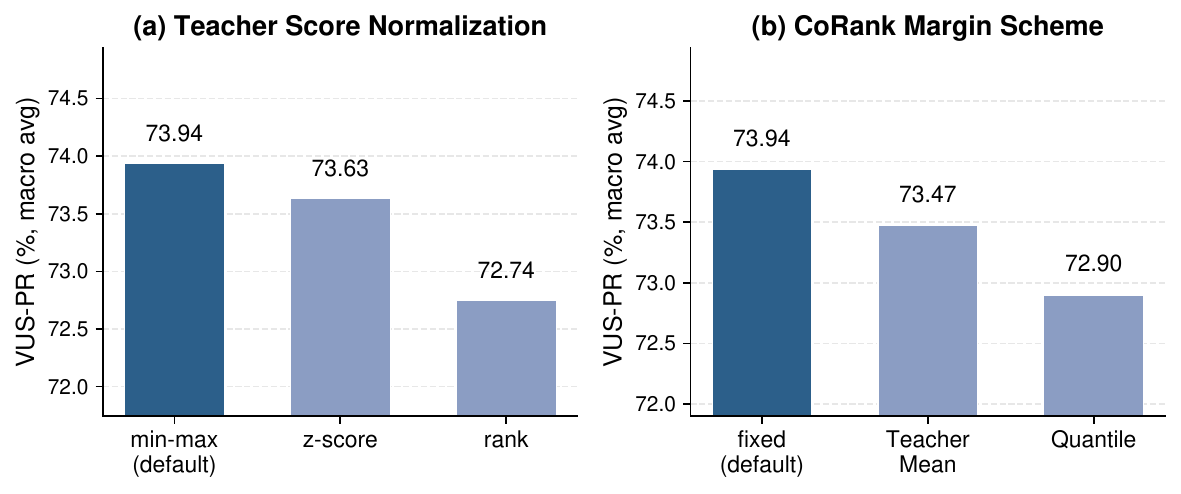}
    \caption{Ablation on teacher score normalization (a) and score-gap threshold selection (b). Each bar is the seven-dataset macro VUS-PR averaged over five seeds. Default settings are colored in dark blue.}
    \label{fig:norm_margin_ablation}
\end{figure}

This appendix reports two additional ablations on design choices that are fixed in the main experiments.
Figure~\ref{fig:norm_margin_ablation}(a) and Figure~\ref{fig:norm_margin_ablation}(b) both report the seven-dataset macro VUS-PR averaged over five seeds.

\paragraph{Teacher Score Normalization.}
Figure~\ref{fig:norm_margin_ablation}(a) compares three ways of normalizing each teacher's training scores before they enter the CoRank objective.
Min-max maps each teacher's training-set range to $[0, 1]$.
Z-score normalizes by the training mean and standard deviation.
Rank converts each score to its empirical CDF position in $[0, 1]$.
Min-max attains the highest macro VUS-PR, with z-score and rank both within roughly one point.

\paragraph{Score-Gap Threshold Selection.}
CoRank uses the score-gap threshold $\delta$ to decide whether a time-point pair is informative enough for training.
In the default setting, a pair $(i,j)$ is retained when at least one teacher has a normalized score gap larger than $\delta=0.1$, namely when $\max_m|\Delta_{ij}^{(m)}|>\delta$.
This shared threshold is simple, but it can be less appropriate if different teachers have very different score-gap distributions.

Figure~\ref{fig:norm_margin_ablation}(b) asks whether replacing the shared threshold with teacher-specific thresholds improves performance.
We compare the default Fixed rule with two alternatives computed from training-set score gaps.
Teacher Mean sets $\delta_m=\mathbb{E}_{i,j}[|\Delta_{ij}^{(m)}|]$, which filters out pairs below teacher $m$'s average separation.
Quantile sets $\delta_m$ to the $25$\textsuperscript{th} percentile of the same gap distribution, so that each teacher filters the same fraction of small-gap pairs.
Both alternatives are parameter-free and use no anomaly labels.

On the seven-dataset macro VUS-PR, the fixed scalar matches or slightly exceeds both per-teacher alternatives.
This suggests that, after min-max normalization, a shared $\delta=0.1$ is sufficient in our setting.
Additional teacher-specific thresholding therefore provides little benefit in this experiment.

\clearpage
\section{Teacher Alignment and Consensus Confidence}
\label{app:teacher_alignment}

We examine where CoRank obtains supervision from the Sel-3 teachers and how teacher agreement relates to student performance.
Teacher alignment is defined as $a_{ij}=(1+c_{ij})/2$, ranging from $50\%$ for balanced opposing score-gap magnitudes to $100\%$ when all nonzero gaps share the same direction.

\paragraph{Training-Pair Distribution.}
We sample two million candidate pairs per dataset and apply the CoRank support rule with $\delta=0.1$.
Table~\ref{tab:training_confidence} reports confidence distributions conditional on retention.
Most retained pairs have high confidence, while SWaT and SMD contain larger shares of conflicting comparisons.

\begin{table}[ht]
\centering
\caption{Confidence distribution among retained training pairs (\%). Rows sum to $100\%$ up to rounding.}
\label{tab:training_confidence}
\small
\begin{tabular}{c|ccccc}
\toprule
Dataset & $c\in[0,0.2)$ & $c\in[0.2,0.4)$ & $c\in[0.4,0.6)$ & $c\in[0.6,0.8)$ & $c\in[0.8,1]$ \\
\midrule
PSM & 0.04 & 0.09 & 0.08 & 0.40 & 99.39 \\
SWaT & 4.06 & 5.16 & 7.36 & 11.65 & 71.76 \\
Exathlon & 0.01 & 0.02 & 0.02 & 0.05 & 99.91 \\
SMD & 3.72 & 4.52 & 6.14 & 9.85 & 75.77 \\
SMAP & 2.30 & 2.02 & 2.71 & 5.61 & 87.36 \\
Daphnet & 0.40 & 0.58 & 0.83 & 2.22 & 95.96 \\
SWAN & 0.18 & 0.41 & 1.24 & 4.33 & 93.84 \\
\bottomrule
\end{tabular}
\end{table}

\paragraph{Confidence-Filtered Training.}
We retrain AnoMix on SWaT with five seeds, keeping the architecture, pair support, optimizer, and training schedule fixed.
For each cutoff, loss terms below that confidence receive zero weight, while the original loss denominator is unchanged.
Full denotes the matched unfiltered run.
Table~\ref{tab:confidence_filtered_training} shows that every cutoff lowers mean VUS-ROC and VUS-PR relative to full CoRank.
This supports retaining comparisons with small but nonzero confidence and weighting them softly.
The VUS-PR trend is not monotonic, and the variation across seeds does not establish a separate contribution for each interval.

\begin{table}[ht]
\centering
\caption{Confidence-filtered training on SWaT (5-seed mean $\pm$ standard deviation, \%).}
\label{tab:confidence_filtered_training}
\small
\begin{tabular}{c|cc}
\toprule
Training confidence range & VUS-ROC & VUS-PR \\
\midrule
$[0,1]$ (Full) & $78.55\pm4.32$ & $42.39\pm3.34$ \\
$[0.2,1]$ & $75.93\pm5.74$ & $39.74\pm4.25$ \\
$[0.4,1]$ & $75.76\pm5.88$ & $41.39\pm4.94$ \\
$[0.6,1]$ & $73.84\pm6.10$ & $39.56\pm3.82$ \\
$[0.8,1]$ & $73.12\pm7.32$ & $40.79\pm5.56$ \\
\bottomrule
\end{tabular}
\end{table}

\noindent\begin{minipage}{\linewidth}
\paragraph{Alignment and Pairwise Accuracy.}
We sample two million anomaly--normal test pairs per dataset, retain pairs under the same score-gap rule, and group them by teacher alignment using training-statistic score normalization.
A preference is correct when the anomalous point receives the higher score, with half credit for ties.
Table~\ref{tab:alignment_pairwise_accuracy} compares the mean individual-teacher accuracy with the seed-$42$ student's accuracy, macro-averaged over the seven datasets.
AnoMix is more accurate in every interval, including those with strongly conflicting teacher gaps.
This is a label-aware post-hoc diagnostic rather than a causal training analysis.
Teacher test scores and test labels are used only for this diagnostic and never to train the student.

\par\medskip
\centering
\captionsetup{hypcap=false}
\captionof{table}{Pairwise accuracy by teacher alignment (seven-dataset macro average, \%).}
\label{tab:alignment_pairwise_accuracy}
\small
\begin{tabular}{c|cc}
\toprule
Teacher alignment & Mean teacher & AnoMix \\
\midrule
$[50,60)\%$ & 58.31 & 84.00 \\
$[60,70)\%$ & 57.44 & 84.77 \\
$[70,80)\%$ & 53.27 & 84.98 \\
$[80,90)\%$ & 51.54 & 87.94 \\
$[90,100]\%$ & 83.28 & 91.38 \\
\bottomrule
\end{tabular}
\end{minipage}

\clearpage
\newpage
\section{Statistical Significance}
\label{app:significance}

This appendix reports variability across the five random seeds $\{42,123,456,789,1024\}$ for Tables~\ref{tab:main}--\ref{tab:zero_shot}.
Standard deviations use the population convention (divisor $5$), matching the experiment summaries, and are reported in \%pt.
An entry of $0.00$ may reflect rounding of a small nonzero deviation.

\begin{table*}[h]
\centering
\caption{Standard deviation across five random seeds (\%) for the main results in Table~\ref{tab:main}.}
\label{tab:std_full}
\scriptsize
\resizebox{\textwidth}{!}{
\setlength{\tabcolsep}{3pt}
\begin{tabular}{c|cc|cc|cc|cc|cc|cc|cc}
\toprule
\multirow{2}{*}{\textbf{Method}} & \multicolumn{2}{c|}{\textbf{PSM}} & \multicolumn{2}{c|}{\textbf{SWaT}} & \multicolumn{2}{c|}{\textbf{Exathlon}} & \multicolumn{2}{c|}{\textbf{SMD}} & \multicolumn{2}{c|}{\textbf{SMAP}} & \multicolumn{2}{c|}{\textbf{Daphnet}} & \multicolumn{2}{c}{\textbf{SWAN}} \\
\cmidrule(lr){2-3}\cmidrule(lr){4-5}\cmidrule(lr){6-7}\cmidrule(lr){8-9}\cmidrule(lr){10-11}\cmidrule(lr){12-13}\cmidrule(lr){14-15}
& \textbf{V-ROC} & \textbf{V-PR} & \textbf{V-ROC} & \textbf{V-PR} & \textbf{V-ROC} & \textbf{V-PR} & \textbf{V-ROC} & \textbf{V-PR} & \textbf{V-ROC} & \textbf{V-PR} & \textbf{V-ROC} & \textbf{V-PR} & \textbf{V-ROC} & \textbf{V-PR} \\
\midrule
Omni~\citep{OmniAnomaly-2019} & 0.00 & 0.00 & 0.00 & 0.00 & 0.00 & 0.00 & 0.00 & 0.00 & 0.00 & 0.00 & 0.00 & 0.00 & 0.00 & 0.00 \\
USAD~\citep{USAD-2020} & 0.89 & 0.57 & 1.31 & 0.75 & 0.00 & 0.04 & 0.34 & 1.31 & 0.00 & 0.00 & 0.84 & 0.38 & 0.20 & 0.29 \\
A.T~\citep{AnomalyTransformer-2021} & 4.04 & 2.59 & 10.16 & 6.40 & 6.57 & 6.97 & 2.63 & 1.03 & 7.06 & 4.02 & 7.63 & 1.77 & 3.05 & 0.96 \\
TranAD~\citep{TranAD-2022} & 0.59 & 0.19 & 1.25 & 0.78 & 0.01 & 0.36 & 0.15 & 0.17 & 11.31 & 3.95 & 0.34 & 0.25 & 0.18 & 0.28 \\
TimesNet~\citep{TimesNet-2022} & 0.69 & 0.58 & 1.01 & 0.89 & 0.11 & 0.84 & 0.65 & 2.05 & 4.50 & 1.85 & 0.15 & 0.59 & 0.46 & 0.90 \\
MOMENT~\citep{MOMENT-2024} & 0.13 & 0.01 & 0.30 & 0.24 & 0.00 & 0.01 & 0.59 & 0.13 & 0.90 & 0.48 & 0.01 & 0.06 & 0.04 & 0.08 \\
\midrule
SARAD~\citep{SARAD-2024} & 0.87 & 0.70 & 1.05 & 0.28 & 0.02 & 0.26 & 0.02 & 0.10 & 0.34 & 0.25 & 0.61 & 0.89 & 0.52 & 0.95 \\
CATCH~\citep{CATCH-2025} & 0.16 & 0.30 & 0.03 & 0.01 & 0.01 & 0.11 & 0.01 & 0.11 & 0.00 & 0.01 & 0.01 & 0.01 & 0.01 & 0.03 \\
CrossAD~\citep{CrossAD-2025} & 0.01 & 0.01 & 0.08 & 0.01 & 0.00 & 0.02 & 0.01 & 0.06 & 0.01 & 0.02 & 0.01 & 0.01 & 0.00 & 0.01 \\
OracleAD~\citep{OracleAD-2025} & 2.43 & 2.09 & 6.69 & 3.50 & 0.34 & 2.97 & 1.65 & 3.76 & 2.32 & 4.53 & 1.59 & 5.60 & 1.46 & 2.92 \\
\midrule
\textbf{AnoMix} & 1.27 & 2.14 & 4.33 & 3.34 & 0.00 & 0.02 & 0.42 & 1.29 & 0.05 & 0.06 & 0.28 & 1.66 & 0.51 & 0.67 \\
\bottomrule
\end{tabular}}
\end{table*}

\begin{table*}[h]
\centering
\caption{Standard deviation across five random seeds (\%) for the score-ensemble comparison in Table~\ref{tab:ensemble}.}
\label{tab:std_ensemble}
\scriptsize
\resizebox{\textwidth}{!}{
\setlength{\tabcolsep}{3pt}
\begin{tabular}{c|cc|cc|cc|cc|cc|cc|cc}
\toprule
\multirow{2}{*}{\textbf{Method}} & \multicolumn{2}{c|}{\textbf{PSM}} & \multicolumn{2}{c|}{\textbf{SWaT}} & \multicolumn{2}{c|}{\textbf{Exathlon}} & \multicolumn{2}{c|}{\textbf{SMD}} & \multicolumn{2}{c|}{\textbf{SMAP}} & \multicolumn{2}{c|}{\textbf{Daphnet}} & \multicolumn{2}{c}{\textbf{SWAN}} \\
\cmidrule(lr){2-3}\cmidrule(lr){4-5}\cmidrule(lr){6-7}\cmidrule(lr){8-9}\cmidrule(lr){10-11}\cmidrule(lr){12-13}\cmidrule(lr){14-15}
& \textbf{V-ROC} & \textbf{V-PR} & \textbf{V-ROC} & \textbf{V-PR} & \textbf{V-ROC} & \textbf{V-PR} & \textbf{V-ROC} & \textbf{V-PR} & \textbf{V-ROC} & \textbf{V-PR} & \textbf{V-ROC} & \textbf{V-PR} & \textbf{V-ROC} & \textbf{V-PR} \\
\midrule
Mean         & 2.66 & 1.59 & 3.54 & 1.29 & 0.13 & 1.76 & 0.28 & 1.24 & 0.58 & 0.14 & 0.88 & 3.92 & 0.32 & 0.30 \\
Max          & 3.91 & 2.46 & 6.60 & 2.80 & 0.18 & 2.74 & 0.49 & 1.42 & 0.41 & 0.08 & 1.46 & 3.67 & 0.52 & 0.14 \\
Median       & 2.01 & 0.86 & 0.68 & 0.31 & 0.21 & 2.49 & 0.34 & 1.38 & 0.82 & 0.41 & 0.38 & 1.38 & 0.89 & 1.85 \\
Rank Fusion  & 1.28 & 0.86 & 2.47 & 1.14 & 0.01 & 0.08 & 0.50 & 2.13 & 0.93 & 1.04 & 0.44 & 2.14 & 0.19 & 0.34 \\
RRF~\citep{RRF-2009} & 1.33 & 1.45 & 2.48 & 1.30 & 0.12 & 1.17 & 0.15 & 1.32 & 1.55 & 1.51 & 0.77 & 2.82 & 0.37 & 0.91 \\
\midrule
\textbf{AnoMix} & 1.27 & 2.14 & 4.33 & 3.34 & 0.00 & 0.02 & 0.42 & 1.29 & 0.05 & 0.06 & 0.28 & 1.66 & 0.51 & 0.67 \\
\bottomrule
\end{tabular}}
\end{table*}

\paragraph{Teacher-Set Variants.}
Table~\ref{tab:std_teacher_sets} reports per-dataset standard deviations for Table~\ref{tab:teacher_sets}, with teacher supervision fixed across student seeds.
SWAN LOO-3 and Sel-3 use the same teachers and student runs.

\begin{table*}[!h]
\centering
\caption{Per-dataset standard deviations (\%pt) for the teacher sets in Table~\ref{tab:teacher_sets}.}
\label{tab:std_teacher_sets}
\scriptsize
\resizebox{\textwidth}{!}{
\setlength{\tabcolsep}{3pt}
\begin{tabular}{c|cc|cc|cc|cc|cc|cc|cc}
\toprule
\multirow{2}{*}{\textbf{Method}} & \multicolumn{2}{c|}{\textbf{PSM}} & \multicolumn{2}{c|}{\textbf{SWaT}} & \multicolumn{2}{c|}{\textbf{Exathlon}} & \multicolumn{2}{c|}{\textbf{SMD}} & \multicolumn{2}{c|}{\textbf{SMAP}} & \multicolumn{2}{c|}{\textbf{Daphnet}} & \multicolumn{2}{c}{\textbf{SWAN}} \\
\cmidrule(lr){2-3}\cmidrule(lr){4-5}\cmidrule(lr){6-7}\cmidrule(lr){8-9}\cmidrule(lr){10-11}\cmidrule(lr){12-13}\cmidrule(lr){14-15}
& \textbf{V-ROC} & \textbf{V-PR} & \textbf{V-ROC} & \textbf{V-PR} & \textbf{V-ROC} & \textbf{V-PR} & \textbf{V-ROC} & \textbf{V-PR} & \textbf{V-ROC} & \textbf{V-PR} & \textbf{V-ROC} & \textbf{V-PR} & \textbf{V-ROC} & \textbf{V-PR} \\
\midrule
LOO-1 & 6.07 & 4.27 & 22.60 & 10.70 & 39.37 & 36.16 & 34.56 & 33.08 & 39.45 & 36.99 & 29.93 & 13.42 & 3.05 & 5.10 \\
LOO-2 & 2.41 & 3.93 & 2.14 & 2.02 & 0.01 & 0.03 & 0.42 & 0.76 & 0.06 & 0.06 & 0.28 & 1.76 & 0.18 & 0.33 \\
LOO-3 & 6.25 & 5.11 & 3.10 & 4.75 & 0.01 & 0.04 & 0.42 & 1.29 & 0.04 & 0.05 & 0.38 & 2.01 & 0.51 & 0.67 \\
LOO-4 & 3.11 & 2.55 & 3.16 & 3.35 & 0.00 & 0.02 & 0.38 & 0.99 & 0.04 & 0.04 & 0.38 & 1.99 & 0.43 & 0.80 \\
LOO-5 & 3.86 & 2.53 & 7.15 & 5.06 & 0.01 & 0.04 & 0.38 & 0.99 & 0.04 & 0.04 & 0.38 & 2.01 & 0.27 & 0.84 \\
Sel-3 & 1.27 & 2.14 & 4.33 & 3.34 & 0.00 & 0.02 & 0.42 & 1.29 & 0.05 & 0.06 & 0.28 & 1.66 & 0.51 & 0.67 \\
All-10 & 2.33 & 1.74 & 3.35 & 2.07 & 0.00 & 0.02 & 0.32 & 1.14 & 0.03 & 0.05 & 0.37 & 2.01 & 0.55 & 0.74 \\
\bottomrule
\end{tabular}}
\end{table*}

\paragraph{Zero-Shot Transfer.}
Table~\ref{tab:std_zero_shot} reports per-dataset standard deviations for AnoMix-ZS in Table~\ref{tab:zero_shot}.
Fixed-checkpoint foundation baselines have no across-training-seed standard deviation.

\begin{table*}[!h]
\centering
\caption{Per-dataset standard deviations (\%pt) for AnoMix-ZS in Table~\ref{tab:zero_shot}.}
\label{tab:std_zero_shot}
\scriptsize
\resizebox{\textwidth}{!}{
\setlength{\tabcolsep}{3pt}
\begin{tabular}{c|cc|cc|cc|cc|cc|cc|cc}
\toprule
\multirow{2}{*}{\textbf{Method}} & \multicolumn{2}{c|}{\textbf{PSM}} & \multicolumn{2}{c|}{\textbf{SWaT}} & \multicolumn{2}{c|}{\textbf{Exathlon}} & \multicolumn{2}{c|}{\textbf{SMD}} & \multicolumn{2}{c|}{\textbf{SMAP}} & \multicolumn{2}{c|}{\textbf{Daphnet}} & \multicolumn{2}{c}{\textbf{SWAN}} \\
\cmidrule(lr){2-3}\cmidrule(lr){4-5}\cmidrule(lr){6-7}\cmidrule(lr){8-9}\cmidrule(lr){10-11}\cmidrule(lr){12-13}\cmidrule(lr){14-15}
& \textbf{V-ROC} & \textbf{V-PR} & \textbf{V-ROC} & \textbf{V-PR} & \textbf{V-ROC} & \textbf{V-PR} & \textbf{V-ROC} & \textbf{V-PR} & \textbf{V-ROC} & \textbf{V-PR} & \textbf{V-ROC} & \textbf{V-PR} & \textbf{V-ROC} & \textbf{V-PR} \\
\midrule
AnoMix-ZS & 1.45 & 1.45 & 1.46 & 2.17 & 0.08 & 1.38 & 0.47 & 4.72 & 0.75 & 2.30 & 1.00 & 4.63 & 0.25 & 0.53 \\

\bottomrule
\end{tabular}}
\end{table*}

\clearpage
\section{Broader Impact}
\label{app:broader_impact}

AnoMix can reduce the practical cost of deploying strong time-series anomaly detection by
replacing multi-model teacher ensembles with a single lightweight student. This is especially
useful in monitoring settings where inference latency, memory footprint, and maintenance cost
matter, such as industrial sensor systems, cyber-physical infrastructure, and edge devices.
This makes the framework relevant to on-device AI for self-diagnosis in robots, satellites,
and manufacturing systems, where local anomaly detection can support earlier fault detection
without continuously running large detector ensembles.
By distilling complementary teacher scoring principles into one deployable model, AnoMix can
make stronger anomaly detection more accessible in environments where running several
detectors in parallel is difficult.

The framework also offers a reusable path for building compact TSAD models from existing
detectors. Many application domains already have specialized unsupervised detectors, each
capturing different abnormality criteria. AnoMix provides a way to consolidate these signals
without requiring anomaly labels, which are often expensive, delayed, or incomplete in real
systems. This can support faster deployment cycles and lower the barrier to applying
multi-detector anomaly detection in domains with limited labeling resources.

As with any anomaly detector, AnoMix can produce false positives and false negatives, and
the impact of these errors depends on the deployment domain. False positives may increase
inspection cost or operator fatigue, while false negatives may delay responses to emerging
faults. These risks should be addressed through domain-specific validation, threshold
calibration, human oversight, and continuous monitoring after deployment. Nevertheless, by
improving deployment efficiency while preserving multiple teacher scoring principles, we
believe that AnoMix can have a positive impact on industry from an application perspective.

\section{Limitations and Future Work}
\label{app:limitations}

In this paper, we proposed AnoMix, a new TSAD framework for label-free multi-teacher
distillation. Although the experimental results support the proposed methodology, several
limitations remain. First, if the teacher pool is weak on the target domain, noisy teacher score
sequences can pass unreliable supervision to the student. If several teachers share a correlated
failure mode, their score gaps may also align on an incorrect pairwise preference, causing
CoRank to assign high confidence to an unreliable consensus. Second, if the training split is
too clean or teacher scores are nearly uniform, only a small number of meaningful score gaps
may be available, limiting pairwise anomaly-preference supervision. Future work could address
these issues through automatic teacher-pool construction, label-free TSAD model-selection
criteria~\citep{Unsupervised_Model_Selection-2022}, diversity-aware teacher selection,
source-data selection strategies, and synthetic anomaly injection that enriches the training
data with stronger pairwise training signals.

\clearpage

\end{document}